%% file: main.tex
\documentclass[10pt,twocolumn,letterpaper]{article}

\usepackage{iccv}
\usepackage{times}
\usepackage{epsfig}
\usepackage{graphicx}
\usepackage{amsmath}
\usepackage{amssymb}

\usepackage[accsupp]{axessibility}

\usepackage[pagebackref=true,breaklinks=true,colorlinks,bookmarks=false]{hyperref}

\usepackage[capitalize]{cleveref}
\crefname{section}{Sec.}{Secs.}
\Crefname{section}{Section}{Sections}
\Crefname{table}{Table}{Tables}
\crefname{table}{Tab.}{Tabs.}

\input{tex/packages.tex}

\input{tex/macros.tex}

\iccvfinalcopy 

\def\iccvPaperID{7224} 

\ificcvfinal\fi

\begin{document}

\title{\vspace{-0.5cm}Semantic Attention Flow Fields for Monocular Dynamic Scene Decomposition}

\author{
Yiqing Liang\hspace{0.5cm}
Eliot Laidlaw\hspace{0.5cm}
Alexander Meyerowitz\hspace{0.5cm}
Srinath Sridhar\hspace{0.5cm}
James Tompkin\\
Brown University
}

\ificcvfinal\thispagestyle{empty}\fi

\twocolumn[{%
\renewcommand\twocolumn[1][]{#1}%
\maketitle
}]

\input{tex/00_abstract}
\input{tex/01_introduction}
\input{tex/02_related}

\input{tex/03_methods}

\input{tex/04_experiments}

\input{tex/05_discussion}

\vspace{-0.5cm}
\paragraph{Acknowledgements} The CV community in New England for feedback, and funding from NSF CNS-2038897 and an Amazon Research Award. Eliot thanks a Randy F.~Pausch '82 Undergraduate Summer Research Award at Brown CS.


{\small
\bibliographystyle{ieee_fullname}
\bibliography{bib/references}
}



\end{document}

%% file: tex/packages.tex
\usepackage{booktabs}       
\usepackage{multirow}
\usepackage{makecell}

\usepackage[table]{xcolor}
\usepackage{amsmath}
\usepackage{amssymb}
\usepackage{graphicx}
\usepackage{siunitx}
\usepackage{xspace}
\usepackage{pifont}
\usepackage{hyperref}
\usepackage{verbatim}
\usepackage{todonotes}
\usepackage{comment}
\usepackage{adjustbox}
\usepackage{algorithm}
\usepackage{algorithmicx}
\usepackage{algpseudocode}
\usepackage{stfloats}

\usepackage{amsfonts}       
\usepackage{nicefrac}       
\usepackage{microtype}      
\usepackage[super]{nth}

\usepackage{enumitem}
\usepackage{lipsum}
\usepackage{wrapfig}
\usepackage{tikz}
\usetikzlibrary{calc,spy}
\usepackage[subtle,title=normal,mathspacing=normal]{savetrees}

\usepackage{doi}
\usepackage{float}
\usepackage{subcaption}

%% file: tex/macros.tex
\newcommand{\isdraft}{false}

\newcommand{\ray}{\mathbf{r}}

\newcommand{\xpos}{\mathbf{x}}

\newcommand{\xcol}{\mathbf{c}}
\newcommand{\pixcol}{\hat{\mathbf{c}}}
\newcommand{\direction}{\boldsymbol{\omega}}

\newcommand{\network}{F_{\boldsymbol\theta}}

\newcommand{\static}{\text{st}}
\newcommand{\dynamic}{\text{dy}}

\newcommand{\timestep}{i}
\newcommand{\flowscene}{\mathbf{f}} 
\newcommand{\flowoptic}{\hat{\mathbf{p}}} 
\newcommand{\semfeat}{\mathbf{s}}
\newcommand{\attfeat}{\mathbf{a}}

\newcommand{\titlename}{SAFF\xspace}
\newcommand{\dataname}{Dynamic Scene Dataset (Masked)\xspace}
\newcommand{\dycheck}{DyCheck Dataset (Masked)\xspace}
\newcommand{\semfeatnetwork}{DINO-ViT\xspace}
\newcommand{\ddnerf}{D$^2$NeRF\xspace}

\newcommand{\loss}{\mathcal{L}}

\newcommand{\lossNSFFPixCol}{\loss_{\pixcol}}

\newcommand{\lossNSFFPixColProj}{\loss_{\pixcol_{i\rightarrow{j}}}}
\newcommand{\lambdaNSFFPixColProj}{\lambda_{\pixcol_{i\rightarrow{j}}}}

\newcommand{\lossNSFFd}{\loss_{\hat{d}}}
\newcommand{\lambdaNSFFd}{\lambda_{\hat{d}}}
\newcommand{\lossNSFFof}{\loss_{\flowoptic}}
\newcommand{\lambdaNSFFof}{\lambda_{\flowoptic}}

\newcommand{\lossSAFF}{\loss_{\text{\titlename}}}
\newcommand{\lossSAFFsem}{\loss_{\hat{\semfeat}}}
\newcommand{\lambdaSAFFsem}{\lambda_{\hat{\semfeat}}}
\newcommand{\lossSAFFatt}{\loss_{\hat{\attfeat}}}
\newcommand{\lambdaSAFFatt}{\lambda_{\hat{\attfeat}}}
\newcommand{\lossSAFFsemproj}{\loss_{\hat{\semfeat}_{i\rightarrow{j}}}}
\newcommand{\lambdaSAFFsemproj}{\lambda_{\hat{\semfeat}_{i\rightarrow{j}}}}
\newcommand{\lossSAFFattproj}{\loss_{\hat{\attfeat}_{i\rightarrow{j}}}}
\newcommand{\lambdaSAFFattproj}{\lambda_{\hat{\attfeat}_{i\rightarrow{j}}}}

\newcommand{\lossSAFFsempyr}{\loss_{\mathcal{P}\hat{\semfeat}}}
\newcommand{\lossSAFFattpyr}{\loss_{\mathcal{P}\hat{\attfeat}}}

\newcommand{\datasplitInput}{Input}
\newcommand{\datasplitCamZero}{Fix Cam 0}
\newcommand{\datasplitTimeZero}{Fix Time 0}
\newcommand{\datasplitTest}{Test}

\newcommand{\seqBalloonNoticeBoard}{\emph{Balloon NBoard}\xspace}
\newcommand{\seqBalloonWall}{\emph{Balloon Wall}\xspace}
\newcommand{\seqDynamicFace}{\emph{Dynamic Face}\xspace}
\newcommand{\seqJumping}{\emph{Jumping}\xspace}

\newcommand{\seqUmbrella}{\emph{Umbrella}\xspace}



%% file: tex/00_abstract.tex
\begin{abstract}
\vspace{-0.15cm}
From video, we reconstruct a neural volume that captures time-varying color, density, scene flow, semantics, and attention information. The semantics and attention let us identify salient foreground objects separately from the background across spacetime. To mitigate low resolution semantic and attention features, we compute pyramids that trade detail with whole-image context. After optimization, we perform a saliency-aware clustering to decompose the scene. To evaluate real-world scenes, we annotate object masks in the NVIDIA Dynamic Scene and DyCheck datasets. We demonstrate that this method can decompose dynamic scenes in an unsupervised way with competitive performance to a supervised method, and that it improves foreground/background segmentation over recent static/dynamic split methods.\\Project webpage: \href{https://visual.cs.brown.edu/saff}{https://visual.cs.brown.edu/saff}
\end{abstract}

%% file: tex/01_introduction.tex
\vspace{-0.5cm}
\section{Introduction}
\vspace{-0.1cm}

Given a casually-captured monocular RGB video of a dynamic scene, decomposing it into foreground objects and background is an important task in computer vision, with downstream applications in segmentation and video editing. An ideal method would also reconstruct the geometry and appearance over time including frame correspondences.

Previous methods have made great progress but there are many limitations. Some works assume that there is no object motion in the scene~\cite{stelzner2021decomposing, yu2022unsupervised, smith2022colf, kobayashi2022distilledfeaturefields, burgess2019monet}, take input from multi-view cameras~\cite{kobayashi2022distilledfeaturefields,yu2022unsupervised}, or do not explicitly reconstruct the underlying 3D structure of the scene~\cite{burgess2019monet, elsayed2022savipp}. For objects, some works rely on masks or user input to aid segmentation~\cite{yang2021objectnerf, kipf2022conditional, elsayed2022savipp}, or use task-specific training datasets~\cite{elsayed2022savipp}. 
Sometimes, works assume the number of foreground objects~\cite{burgess2019monet, locatello2020object}. Given the challenges, many works train and test on synthetic data~\cite{Kabra2021SIMONeVT, pmlr-v100-veerapaneni20a}.

\input{fig/fig_teaser/fig_teaser_final}

We present Semantic Attention Flow Fields (\titlename): A method to overcome these limitations by integrating low-level reconstruction cues with high-level pretrained information---both bottom-up and top-down---into a neural volume.
With this, we demonstrate that embedded semantic and saliency (attention) information is useful for unsupervised dynamic scene decomposition.
\titlename builds upon neural scene flow fields~\cite{li2020neural}, an approach that reconstructs appearance, geometry, and motion. This uses frame interpolation rather than explicit canonicalization~\cite{tretschk2020nonrigid} or a latent hyperspace~\cite{park2021hypernerf}, which lets it more easily apply to casual videos. For optimization, we supervise two network heads with pretrained \semfeatnetwork~\cite{caron2021emerging} semantic features and attention. Naively supervising high-resolution \titlename with low-resolution \semfeatnetwork output reduces reconstruction quality. To mitigate the mismatch, we build a semantic attention pyramid that trades detail with whole-image context. Having optimized a \titlename representation for a dynamic scene, we perform a saliency-aware clustering both in 3D and on rendered feature images to describe objects and their background. Given the volume reconstruction, the clustering generalizes to novel spacetime views.

To evaluate \titlename's dynamic scene decomposition capacity, we expand the NVIDIA Dynamic Scene~\cite{Yoon_2020_CVPR} and DyCheck~\cite{gao2022dynamic} datasets by manually annotating object masks across input and hold-out views. We demonstrate that \titlename outperforms 2D \semfeatnetwork baselines and is comparable to a state-of-the-art video segmentation method ProposeReduce\cite{lin2021video} on our data. Existing monocular video dynamic volume reconstruction methods typically separate static and dynamic parts, but these often do not represent meaningful foreground. We show improved foreground segmentation over NSFF and 
the current D$^2$NeRF~\cite{wu2022ddnerf} method for downstream tasks like editing.


%% file: fig/fig_teaser/fig_teaser_final.tex
{
\begin{figure}[t]
    \centering
    \vspace{-0.35cm}
    \includegraphics[width=0.97\linewidth]{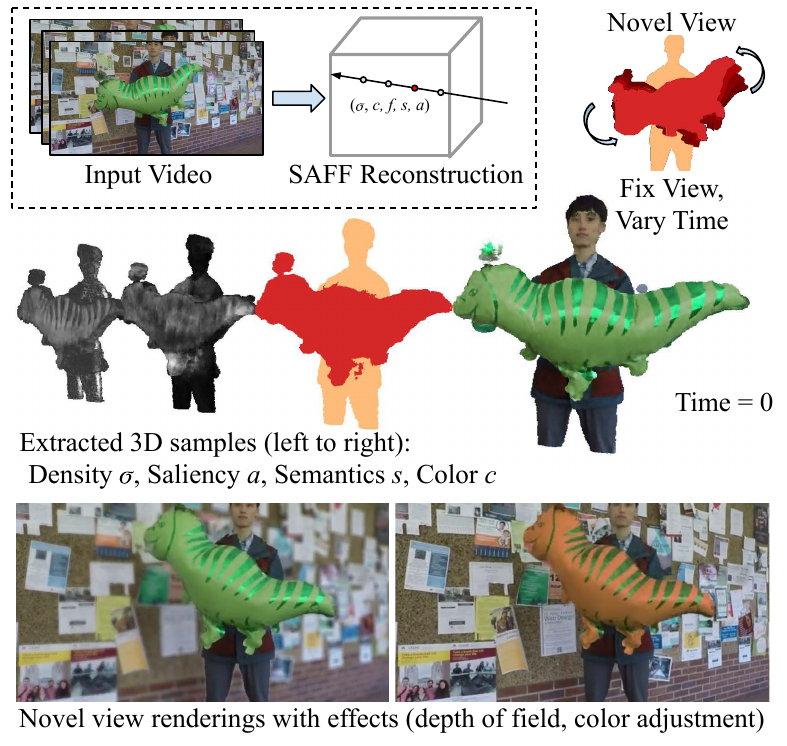}
    \vspace{-0.2cm}
    \caption{
        We decompose a dynamic 3D scene from monocular input into time-varying color, density, scene flow, semantics, and attention information. This could be used to provide precise editing of objects or effects focused around objects.
    }
    \vspace{-0.35cm}
    
\end{figure}
}

%% file: tex/02_related.tex

\section{Related Work}
\label{sec:relatedwork}

\vspace{-0.1cm}

Decomposing a scene into regions of interest is a long-studied task in computer vision~\cite{Caelles_arXiv_2019}, including class, instance, and panoptic segmentation in high-level or top-down vision, and use of low-level or bottom-up cues like motion. Recent progress has considered images~\cite{greff2019multiobject,burgess2019monet,locatello2020object}, videos~\cite{Kabra2021SIMONeVT,kipf2022conditional}, layer decomposition~\cite{ye2022sprites,Monnier_2021_ICCV}, and in-the-wild databases using deep generative models~\cite{Niemeyer2020GIRAFFE}. 
One example, SAVi++\cite{elsayed2022savipp}, uses slot attention~\cite{locatello2020object} to define 2D objects in real-world videos.
Providing first-frame segmentation masks achieves stable performance, with validation on the driving Waymo Open Dataset~\cite{Sun_2020_CVPR}. 
Our work attempts 3D scene decomposition for a casual monocular video without initial masks.

\vspace{-0.14cm}
\paragraph{Scene Decomposition with NeRFs}
\label{sec:relatedwork_withnerfs}

Neural Radiance Fields (NeRF)~\cite{10.1111:cgf.14505} have spurred new scene decomposition research through volumes.
ObSuRF~\cite{stelzner2021decomposing} and uORF~\cite{yu2022unsupervised} are unsupervised slot attention works that bind a latent code to each object. 
Unsupervised decomposition is also possible on light fields~\cite{smith2022colf}. 
For dynamic scenes, works like NeuralDiff~\cite{tschernezki21neuraldiff} and \ddnerf~\cite{wu2022ddnerf} focus on foreground separation, where foreground is defined to contain moving objects. 
Other works like N3F~\cite{tschernezki22neural} and occlusions-4d~\cite{vanhoorick2022revealing} also decompose foregrounds into individual objects. 
N3F requires user input to specify which object to segment, and occlusions-4d takes RGB point clouds as input. 
Our work attempts to recover a segmented dynamic scene from a monocular RGB video without added markup or masks.

\vspace{-0.14cm}
\paragraph{Neural Fields Beyond Radiance}
\label{sec:relatedwork_beyondradiance}

Research has begun to add non-color information to neural volumes to aid decomposition via additional feature heads on the MLP. 
iLabel~\cite{Zhi:etal:ICCV2021} adds a semantic head to propagate user-provided segmentations in the volume. PNF~\cite{Kundu_2022_CVPR} and Panoptic-NeRF~\cite{fu2022panoptic} attempt panoptic segmentation within neural fields, and Object-NeRF integrates instance segmentation masks into the field during optimization~\cite{yang2021objectnerf}. 
Research also investigates how to apply generic pretrained features to neural fields, like \semfeatnetwork.

\vspace{-0.14cm}
\paragraph{\semfeatnetwork for Semantics and Saliency}
\semfeatnetwork is a self-supervised transformer that, after pretraining, extracts generic semantic information~\cite{caron2021emerging}. Amir et al.~\cite{amir2021deep} use \semfeatnetwork features with $k$-means clustering to achieve co-segmentation across a video. Seitzer et al.~\cite{seitzer2022bridging} combine slot attention and \semfeatnetwork features for object-centric learning on real-world 2D data. TokenCut~\cite{wang2022tokencut} performs normalized cuts on \semfeatnetwork features for foreground segmentation on natural images. Deep Spectral Segmentation~\cite{melaskyriazi2022deep} show that graph Laplacian processing of \semfeatnetwork features provides unsupervised foreground segmentation, and Selfmask~ \cite{shin2022selfmask} shows that these features can provide object saliency masks. 
Our approach considers these clustering and saliency findings for the setting of 3D decomposition from monocular video.

\vspace{-0.14cm}
\paragraph{\semfeatnetwork Fields}
\label{sec:concurrent}

Concurrent works have integrated \semfeatnetwork features into neural fields. DFF~\cite{kobayashi2022distilledfeaturefields} distills features for dense multi-view static scenes with user input for segmentation.
N3F~\cite{tschernezki22neural} expands NeuralDiff to dynamic scenes, and relies on user input for segmentation. AutoLabel~\cite{blomqvist2022baking} uses \semfeatnetwork features to accelerate segmentation in static scenes given a ground truth segmentation mask.
%
Other works use \semfeatnetwork differently. FeatureRealisticFusion~\cite{Mazur:etal:ARXIV2022} uses \semfeatnetwork in an online feature fusion task, focusing on propagating user input, and 
NeRF-SOS~\cite{Fan2022} uses 2D DINO-ViT to process NeRF-rendered multi-view RGB images of a static scene.
%
In contrast to these works, we consider real-world casual monocular videos, recover and decompose a 3D scene, then explore whether saliency can avoid the need for masks or user input in segmenting objects.

%
\input{fig/fig_overview/fig_overview.tex}

%% file: fig/fig_overview/fig_overview.tex

\begin{figure*}
    \centering
    \includegraphics[width=0.97\linewidth]{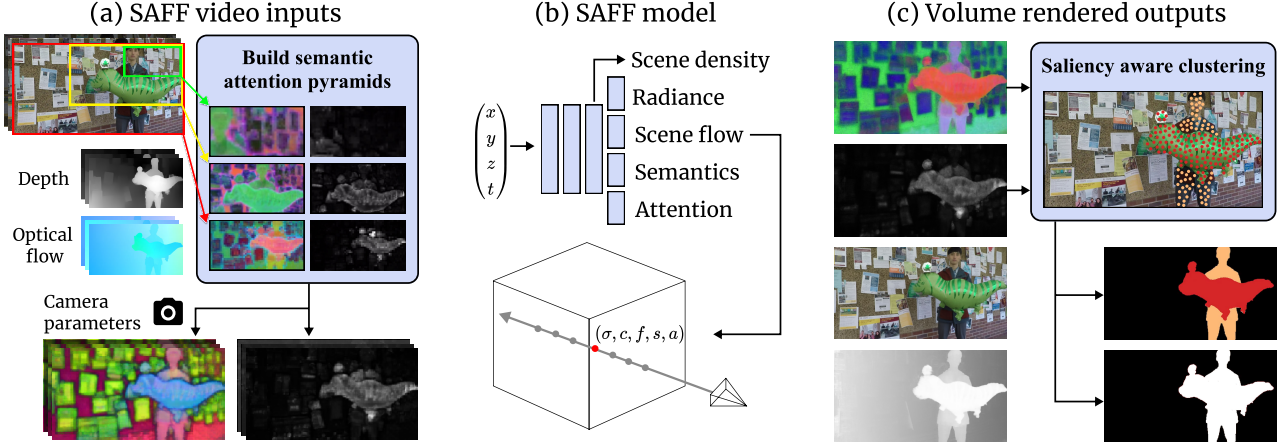}
    \vspace{-0.4cm}
    \caption{
        \textbf{Overview.} From monocular video, \titlename builds a neural field of scene-flowed 3D density, radiance, semantics, and attention (b). This is guided by semantic attention pyramids that increase resolution, plus depth and optical flow priors (a). We can render new spacetime views of any channel, and use saliency-aware clustering to decompose objects and background (c).
    }
    \label{fig:pipeline}
    \vspace{-0.35cm}
\end{figure*}

%% file: tex/03_methods.tex
\section{Method}
\label{sec:method}

\vspace{-0.15cm}
For a baseline dynamic scene reconstruction method, we begin with NSFF from Li \etal~\cite{li2020neural} (\cref{sec:method_basic}), which builds upon NeRF~\cite{mildenhall2020nerf}. NSFF's low-level scene flow frame-to-frame approach provides better reconstructions for real-world casual monocular videos than deformation-based methods~\cite{tretschk2020nonrigid,park2021hypernerf}. We modify the architecture to integrate higher-level semantic and saliency (or attention) features (\cref{sec:method_semanticsalient}). 
After optimizing a \titlename for each scene, we perform saliency-aware clustering of the field (\cref{sec:method_saliencyawareclustering}). All implementation details are in our supplemental material.

\vspace{-0.15cm}
\paragraph{Input} Our method takes in a single RGB video over time $\timestep$ as an ordered set of images $I \in \mathcal{I}$ and camera poses. We use COLMAP to recover camera poses~\cite{schoenberger2016sfm}. From all poses, we define an NDC-like space that bounds the scene, and a set of rays $\ray \in \mathcal{R}$, one per image pixel with color $\hat{\xcol}^\dagger$. Here, $\hat{\cdot}$ denotes a 2D pixel value in contrast to a 3D field value, and $\cdot^\dagger$ denotes an input value in contrast to an estimated value.

From pretrained networks, we estimate single-frame monocular depth $\hat{d}^\dagger$ (MiDaSv2~\cite{Ranftl2022}), optical flow $\flowoptic_{i}^\dagger$ (RAFT~\cite{2020RAFT}), and semantic features $\hat{\semfeat}^\dagger$ and attention $\hat{\attfeat}^\dagger$ (\semfeatnetwork~\cite{caron2021emerging}) after important preprocessing (\cref{sec:method_semfeatnetworkprocess}).

\input{tex/03a_methods_nsff.tex}
\input{tex/03b_methods_saff.tex}
\input{tex/03c_methods_saffpyramid.tex}
\input{tex/03d_methods_saffcluster.tex}

%% file: tex/03a_methods_nsff.tex
\subsection{Initial Dynamic Neural Volume Representation}
\label{sec:method_basic}

\vspace{-0.1cm}
The initial representation comprises a static NeRF $\network^\static$ and a dynamic NeRF $\network^\dynamic$.
The static network predicts a color $\xcol$, density $\sigma$, and blending weight $v$ (\cref{eq:nsff_static}), and the dynamic network predicts time-varying color $\xcol_i$, density $\sigma_i$, scene flow $\flowscene_i$, and occlusion weights $w_i$ (\cref{eq:nsff_dynamic}). In both network architectures, other than position $\xpos$, we add direction $\direction$ to a late separate head to only condition the estimation of color $\xcol$.
\begin{align}
\label{eq:nsff_static}
    \network^\static : (\xpos, \direction) \hspace{0.3cm} &\rightarrow (\xcol^\static, \sigma^\static, v) \\
\label{eq:nsff_dynamic}
    \network^\dynamic : (\xpos, \direction, \timestep) &\rightarrow (\xcol_i^\dynamic, \sigma_i^\dynamic, \flowscene_i, w_i)
\end{align}
To produce a pixel's color, we sample points at distances $t$ along the ray $\xpos_t=\xpos-\direction t$ between near and far planes $t_n$ to $t_f$, query each network, then integrate transmittance $T$, density $\sigma$, and color along the ray from these samples~\cite{mildenhall2020nerf}.
For brevity, we omit evaluating at $\xpos_t,\direction$, \eg, $\sigma^\static(\xpos_t)$ is simply $\sigma^\static$; $\xcol^\static(\xpos_t,\direction)$ is simply $\xcol^\static$.

We produce a combined color from the static and dynamic colors by multiplication with their densities:
\vspace{-0.1cm}
\begin{align}
\label{eq:nsff_color_staticdynamic_sum}
    \sigma_i \xcol_i = v\sigma^\static\xcol^\static + (1-v)\sigma_i^\dynamic\xcol_i^\dynamic
    \vspace{-0.1cm}
\end{align}
Given that transmittance integrates density up to the current sampled point under Beer-Lambert volume attenuation, the rendered pixel color for the ray is computed as:
\vspace{-0.1cm}
\begin{align}
\label{eq:nsff_colorintegral}
    \pixcol_i =  \int_{t_n}^{t_f} {T_i \sigma_i \xcol_i \, dt} \hspace{0.2cm}\text{where}\hspace{0.2cm} T_i = \exp\left( - \int_{t_n}^{t} \sigma_i \, dt\right)
\end{align}
To optimize the volume to reconstruct input images, we compute a photometric loss $L_{\hat{\xcol}}$ between rendered and input colors: 
\begin{align}
\label{eq:nsff_loss_color}
    \lossNSFFPixCol = \frac{1}{|\mathcal{R}|} \sum\limits_{\ray_i \in \mathcal{R}}||\pixcol_i(\ray_i) - \pixcol_i^\dagger(\ray_i)||_2^2
\end{align}
%
\vspace{-0.6cm}
\paragraph{Scene Flow} Defining correspondence over time is important for monocular input as it lets us penalize a reprojection error with neighboring frames $j \in \mathcal{N}$, e.g., where $j=i+1$ or $j=i-1$. We denote $_{i\rightarrow{j}}$ for the projection of frame $i$ onto frame $j$ by scene flow. $\network^\dynamic$ estimates both forwards and backwards scene flow at every point to penalize a bi-direction loss.

Thus, we approximate color output at time step $j$ by flowing the queried network values at time step $i$:%
\vspace{-0.05cm}
\begin{align}
\label{eq:nsff_loss_color_reprojection}
    \pixcol_{i\rightarrow{j}} = \int_{t_n}^{t_f} T_{i\rightarrow j}\sigma_{i\rightarrow j}\xcol_{i\rightarrow j} \, dt
\end{align}
Reprojection must account for occlusion and disocclusion by motion. As such, $\network^\dynamic$ also predicts forwards and backwards scene occlusion weights $w_{i+1}$ and $w_{i-1}\in [0, 1]$, where a point with $w_{i+1}=0$ means that occlusion status is changed one step forwards in time. We can integrate $w$ to a pixel:
\vspace{-0.05cm}
\begin{align}
\label{eq:nsff_weight_projection}
    \hat{w}_{i\rightarrow{j}} = \int_{t_n}^{t_f} T_{i\rightarrow{j}}\sigma_{i\rightarrow{j}}w_j \, dt 
\end{align}
Then, this pixel weight modulates the color loss such that occluded pixels are ignored. 
\vspace{-0.05cm}
\begin{align}
\label{eq:nsff_loss_photometric}
    \lossNSFFPixColProj = \frac{1}{|\mathcal{R}||\mathcal{N}|} \sum\limits_{\ray_i \in \mathcal{R}}\sum\limits_{j\in \mathcal{N}} \hat{w}_{i\rightarrow{j}}(\ray_i) &||\pixcol_{i\rightarrow{j}}(\ray_i) - \pixcol_j^\dagger(\ray_j)||_2^2
    \raisetag{3.5\normalbaselineskip} 
\end{align}
%
%
%
%
\paragraph{Prior losses}
We use the pretrained depth and optical flow map losses to help overcome the ill-posed monocular reconstruction problem. These losses decay as optimization progresses to rely more and more on the optimized self-consistent geometry and scene flow. 
For geometry, we estimate a depth $\hat{d}_i$ for each ray $\ray_i$ by replacing $\xcol_i$ in \cref{eq:nsff_colorintegral} by the distance $t$ along the ray. Transform $z$ estimates a scale and shift as the pretrained network produces only relative depth.
\vspace{-0.1cm}
\begin{align}
\label{eq:nsff_loss_d}
    \lossNSFFd = \frac{1}{|\mathcal{R}|} \sum\limits_{\ray_i \in \mathcal{R}} ||\hat{d}_i - z(\hat{d}_i^\dagger)||_1
\end{align}
For motion, projecting scene flow to a camera lets us compare to the estimated optical flow. Each sample point along a ray $\xpos_i$ is advected to a point in the neighboring frame $\xpos_{i\rightarrow{j}}$, then integrated to the neighboring camera plane to produce a 2D point offset $\flowoptic_{i}(\ray_i)$. Then, we expect the difference in the start and end positions to match the prior: 
%
\vspace{-0.1cm}
\begin{align}
\label{eq:nsff_loss_opticalflow}
    \lossNSFFof = \frac{1}{|\mathcal{R}||\mathcal{N}|} \sum\limits_{\ray_i \in \mathcal{R}}\sum\limits_{j\in \mathcal{N}(i)} ||\flowoptic_{i}(\ray_i) - \flowoptic^\dagger_{i}(\ray_i)||_1
\end{align}

%
%
Additional regularizations encourage occlusion weights to be close to one, scene flow to be small, locally constant, and cyclically consistent, and blending weight $v$ to be sparse.

%% file: tex/03b_methods_saff.tex
\subsection{Semantic Attention Flow Fields}
\label{sec:method_semanticsalient}

Beyond low-level or \emph{bottom-up} features, high-level or \emph{top-down} features are also useful to define objects and help down-stream tasks like segmentation. For example, methods like NSFF or D$^2$NeRF struggle to provide \emph{useful} separation of static and dynamic parts because blend weight $v$ estimates whether the volume \emph{appears} to be occupied by some moving entity. This is not the same as objectness; tasks like video editing could benefit from accurate dynamic object masks.


As such, we extract 2D semantic features and attention (or saliency) values from a pretrained \semfeatnetwork network, then optimize the \titlename such that unknown 3D semantic and attention features over time can be projected to recreate their 2D complements.
This helps us to ascribe semantic meaning to the volume and to identify objects.
As semantics/attention are integrated into the 4D volume, we can render them from novel spacetime views without further DINO-ViT computation. 

To estimate semantic features $\semfeat$ and attention $\attfeat$ at 3D points in the volume at time $\timestep$, we add two new heads to both the static $\network^\static$ and the dynamic $\network^\dynamic$ networks:
%
\begin{align}
\label{eq:saff_static}
    \network^\static : (\xpos, \direction) \hspace{0.3cm} &\rightarrow (\xcol^\static, \sigma^\static, v, \semfeat^\static, \attfeat^\static) \\
\label{eq:saff_dynamic}
    \network^\dynamic : (\xpos, \direction, \timestep) &\rightarrow (\xcol_i^\dynamic, \sigma_i^\dynamic, \flowscene_i, w_i, \semfeat_i^\dynamic, \attfeat_i^\dynamic)
\end{align}
%
As semantic features have been demonstrated to be somewhat robust to view dependence~\cite{amir2021deep}, in our architectures both heads for $\semfeat,\attfeat$ are taken off the backbone before $\direction$ is injected.

To render semantics from the volume, we replace the color term $\xcol$ in \cref{eq:nsff_colorintegral} with $\semfeat$ and equivalently for $\attfeat$:%
%
\vspace{-0.1cm}
\begin{align}
\label{eq:saff_semfeat_attfeat_blending_integral}
    \sigma_i\semfeat_i &= v\sigma^\static\semfeat^\static + (1-v)\sigma_i^\dynamic\semfeat_i^\dynamic \mathrm{,\hspace{0.1cm}} 
    \hat{\semfeat}_i = \int_{t_n}^{t_f} T_i\sigma_i\semfeat_i \, dt   
\end{align}
%
To encourage scene flow to respect semantics over time, we penalize complementary losses on $\semfeat$ and $\attfeat$ (showing $\semfeat$ only):
\vspace{-0.1cm}
\begin{align}
\label{eq:saff_losses_semfeat_attfeat_flow}
    \lossSAFFsemproj = \frac{1}{|\mathcal{R}||\mathcal{N}|} \sum\limits_{\ray_i \in \mathcal{R}}\sum\limits_{j\in \mathcal{N}} \hat{w}_{i\rightarrow{j}}(\ray_i) ||\hat{\semfeat}_{i\rightarrow{j}}(\ray_i) - \hat{\semfeat}_j^\dagger(\ray_j)||_2^2 
    \raisetag{0.5\normalbaselineskip} 
    %
\end{align}
%
%
Finally, as supervision, we add respective losses on the reconstruction of the 2D semantic and attention features from projected 3D volume points (showing $\semfeat$ only):%
\vspace{-0.1cm}
\begin{align}
\label{eq:saff_losses_semfeat_attfeat}
    \lossSAFFsem &= \frac{1}{|\mathcal{R}|} \sum\limits_{\ray_i \in \mathcal{R}} ||\hat{\semfeat}_i(\ray_i) - \hat{\semfeat}_i^\dagger(\ray_i)||_2^2 
\end{align}
Unlike depth and scene flow priors, these are not priors---there is no self-consistency for semantics to constrain their values. Thus, we \emph{do not} decay their contribution. While decaying avoids disagreements between semantic and attention features and color-enforced scene geometry, it also leads to a loss of useful meaning (please see supplemental). 

Thus our final loss becomes:
\vspace{-0.1cm}
\begin{align}
\label{eq:saff_loss_final}
    \lossSAFF = \lossNSFFPixCol 
    + \lambdaNSFFPixColProj \lossNSFFPixColProj
    + \lambdaNSFFd      \lossNSFFd
    + \lambdaNSFFof     \lossNSFFof \\
    + \lambdaSAFFsemproj \lossSAFFsemproj 
    + \lambdaSAFFattproj \lossSAFFattproj \nonumber 
    + \lambdaSAFFsem \lossSAFFsem + \lambdaSAFFatt \lossSAFFatt 
\end{align}

%
\input{fig/fig_feature_improvement/fig_feature_improvement.tex}

%% file: fig/fig_feature_improvement/fig_feature_improvement.tex
%
%

{   

\begin{figure*}[t]
    \centering
    \setlength{\tabcolsep}{1pt}
    
    \newcommand{\figSeqTwoDim}{fig/fig_feature_improvement/twodim}
    \newcommand{\figSeqVolume}{fig/fig_feature_improvement/volume}

    \newcommand{\graymap}{{0 1}}
    \newcommand{\graymapBrighter}{{0.2 1.2}}
    \newcommand{\rgbcolormap}{{0 1 0 1 0 1}}
    \newcommand{\rgbcolormapFlip}{{0 1 0 1 0 1}}
    \newcommand{\rgbcolormapBrighter}{{1 0 1 0 1 0}}

    \newcommand{\imgw}{0.18\linewidth} 
    \newcommand{\imgwhalf}{0.095\linewidth} 
    
    \newcommand{\imgRGB}[2][0px 0px 0px 0px]{\includegraphics[draft=\isdraft,width=\imgw,trim=#1,clip]{#2}}
    
    \newcommand{\imgAtt}[2][0px 0px 0px 0px]{\includegraphics[draft=\isdraft,decodearray={0.2 1.2},width=\imgw,trim=#1,clip]{#2}}

    \newcommand{\imgAttBr}[2][0px 0px 0px 0px]{\includegraphics[draft=\isdraft,decodearray={0.2 1.4},width=\imgw,trim=#1,clip]{#2}}


    \newcommand{\zoomRGB}[4]{
        \begin{tikzpicture}[
    		image/.style={inner sep=0pt, outer sep=0pt},
    		collabel/.style={above=9pt, anchor=north, inner ysep=0pt, align=center}, 
    		rowlabel/.style={left=9pt, rotate=90, anchor=north, inner ysep=0pt, scale=0.8, align=center},
    		subcaption/.style={inner xsep=0.75mm, inner ysep=0.75mm, below right},
    		arrow/.style={-{Latex[length=2.5mm,width=4mm]}, line width=2mm},
    		spy using outlines={rectangle, size=1.25cm, magnification=3, connect spies, ultra thick, every spy on node/.append style={thick}},
    		style1/.style={cyan!90!black,thick},
    		style2/.style={orange!90!black},
    		style3/.style={blue!90!black},
    		style4/.style={green!90!black},
    		style5/.style={white},
    		style6/.style={black},
        ]
        
        \node [image] (#1) {\imgRGB[100px 0px 0px 0px]{#2}};
        \spy[style5] on ($(#1.center)-#3$) in node (crop-#1) [anchor=#4] at (#1.#4);
        
        \end{tikzpicture}
    }

    \newcommand{\zoomAtt}[4]{
        \begin{tikzpicture}[
    		image/.style={inner sep=0pt, outer sep=0pt},
    		collabel/.style={above=9pt, anchor=north, inner ysep=0pt, align=center}, 
    		rowlabel/.style={left=9pt, rotate=90, anchor=north, inner ysep=0pt, scale=0.8, align=center},
    		subcaption/.style={inner xsep=0.75mm, inner ysep=0.75mm, below right},
    		arrow/.style={-{Latex[length=2.5mm,width=4mm]}, line width=2mm},
    		spy using outlines={rectangle, size=1.25cm, magnification=3, connect spies, ultra thick, every spy on node/.append style={thick}},
    		style1/.style={cyan!90!black,thick},
    		style2/.style={orange!90!black},
    		style3/.style={blue!90!black},
    		style4/.style={green!90!black},
    		style5/.style={white},
    		style6/.style={black},
        ]
        
        \node [image] (#1) {\imgAtt[100px 0px 0px 0px]{#2}};
        \spy[style5] on ($(#1.center)-#3$) in node (crop-#1) [anchor=#4] at (#1.#4);
        
        \end{tikzpicture}
    }

    \newcommand{\zoomAttBr}[4]{
        \begin{tikzpicture}[
    		image/.style={inner sep=0pt, outer sep=0pt},
    		collabel/.style={above=9pt, anchor=north, inner ysep=0pt, align=center}, 
    		rowlabel/.style={left=9pt, rotate=90, anchor=north, inner ysep=0pt, scale=0.8, align=center},
    		subcaption/.style={inner xsep=0.75mm, inner ysep=0.75mm, below right},
    		arrow/.style={-{Latex[length=2.5mm,width=4mm]}, line width=2mm},
    		spy using outlines={rectangle, size=1.25cm, magnification=3, connect spies, ultra thick, every spy on node/.append style={thick}},
    		style1/.style={cyan!90!black,thick},
    		style2/.style={orange!90!black},
    		style3/.style={blue!90!black},
    		style4/.style={green!90!black},
    		style5/.style={white},
    		style6/.style={black},
        ]
        
        \node [image] (#1) {\imgAttBr[100px 0px 0px 0px]{#2}};
        \spy[style5] on ($(#1.center)-#3$) in node (crop-#1) [anchor=#4] at (#1.#4);
        
        \end{tikzpicture}
    }

    \scriptsize
    \begin{tabular}{c c@{\hspace{0.2mm}} c@{\hspace{0.2mm}} c@{\hspace{0.2mm}} c@{\hspace{0.2mm}}  c@{\hspace{0.2mm}} c}
    


    (a) Input &
    (b) Semantics &
    (c) Semantics (volume) & 
    (d) Saliency &
    (e) Saliency (volume) 
    \\    

    %
    \imgRGB[100px 0px 0px 0px]{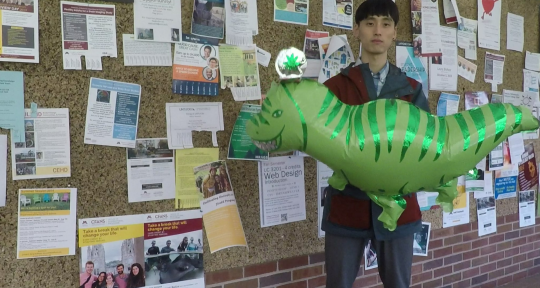} &
     
    \zoomRGB{dino-semantics}{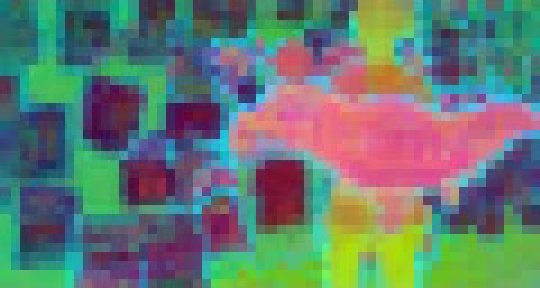}{(-0.1,-0.3)}{south west} &

    \zoomRGB{dino-semantics-volume}{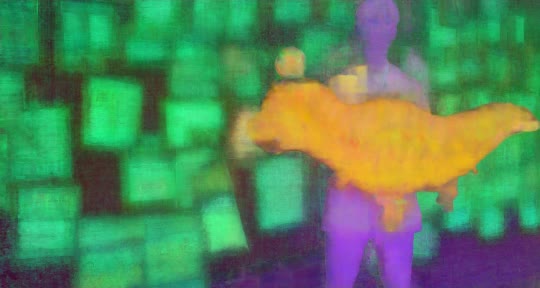}{(-0.1,-0.3)}{south west} &

    \zoomAttBr{dino-saliency}{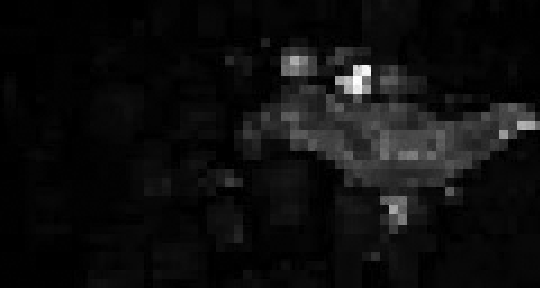}{(-0.1,-0.3)}{south west} &
    
    \zoomAtt{dino-saliency-volume}{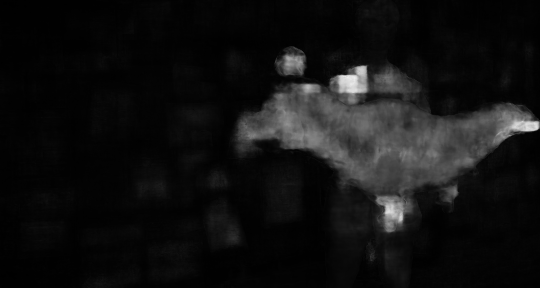}{(-0.1,-0.3)}{south west} \\

    
    &
    (f) Pyr.~semantics &
    (g) Our pyr.~semantics (volume) & 
    (h) Pyr.~saliency &
    (i) Our pyr.~saliency (volume)
     \\

    &
    \zoomRGB{dino-semantics-pyramid}{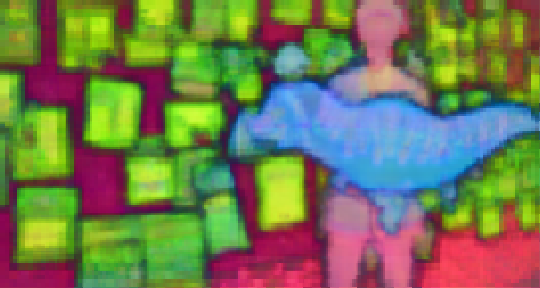}{(-0.1,-0.3)}{south west} &
    
    \zoomRGB{dino-semantics-pyramid-volume}{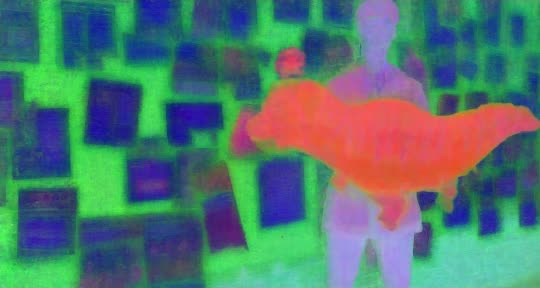}{(-0.1,-0.3)}{south west} &
    
    \zoomAtt{dino-saliency-pyramid}{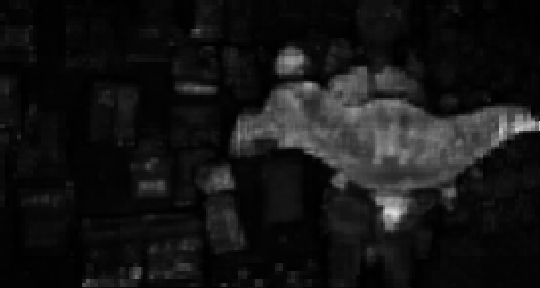}{(-0.1,-0.3)}{south west} &
    
    \zoomAttBr{dino-saliency-pyramid-volume}{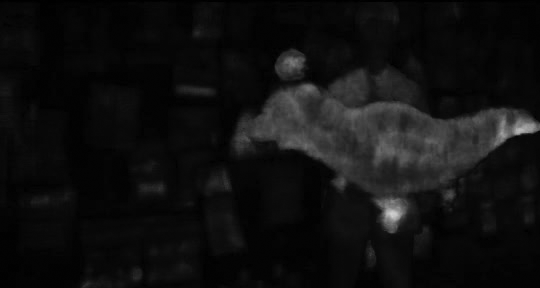}{(-0.1,-0.3)}{south west} &

    \end{tabular}

    \vspace{-0.25cm}
    \caption{
        \textbf{Semantics and saliency improve by both volume integration and by our pyramid.} On \seqBalloonNoticeBoard, blocky artifacts are removed. Semantics are visualized as most significant three PCA dimensions; specific colors are less meaningful.
    }
	
    \label{fig:feature_improvement}
    \vspace{-0.25cm}
\end{figure*}
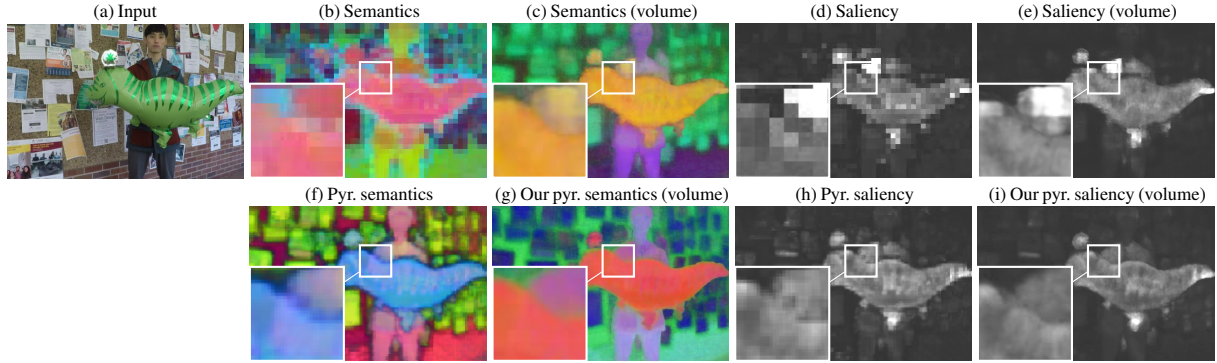

}

%% file: tex/03c_methods_saffpyramid.tex
\vspace{-0.25cm}
\subsection{Semantic Attention Pyramids}
\label{sec:method_semfeatnetworkprocess}






When thinking about scenes, we might argue that semantics from an ideal extractor should be scale invariant, as distant objects have the same class as close objects. 
We might also argue that saliency (or attention features) may not be scale invariant, as small details in a scene should only be salient when viewed close up.
In practice, both extracted features vary across scale and have limited resolution, e.g., \semfeatnetwork~\cite{caron2021emerging} produces one output for each $8\times 8$ patch. 
But, from this, we want semantic features and saliency for every RGB pixel that still respects scene boundaries.

Thus far, work on \emph{static} scenes has ignored the input/feature resolution mismatch~\cite{kobayashi2022distilledfeaturefields} 
as multi-view constraints provide improved localization within the volume. 
For monocular video, this approach has limitations \cite{tschernezki22neural}. Forming many constraints on dynamic objects requires long-term motion correspondence---a tricky task---and so we want to maximize the resolution of any input features where possible without changing their meaning.

One way may be through a pyramid of semantic and attention features that uses a sliding window approach at finer resolutions. Averaging features could increase detail around edges, but we must overcome the practical limit that these features are not stable across scales. This is especially important for saliency: unlike typical RGB pyramids that must preserve energy in an alias-free way~\cite{barron2021mip}, saliency changes significantly over scales and does not preserve energy.

Consider a feature pyramid $\mathcal{P}$ with loss weights per level: 
\vspace{-0.1cm}
\begin{align}
    \lossSAFFsempyr = \sum\limits_{i \in \mathcal{P}} \lambdaSAFFsem^i\lossSAFFsem^i \hspace{0.5cm} \lossSAFFattpyr = \sum\limits_{i \in \mathcal{P}} \lambdaSAFFatt^i\lossSAFFatt^i
\label{eq:weighted_sum_pyr}
\end{align}
Naively encouraging scale-consistent semantics and whole-image saliency, e.g., $\lambdaSAFFsem\! =\! \{\nicefrac{1}{3},\nicefrac{1}{3},\nicefrac{1}{3} \}$ with $\lambdaSAFFatt\! =\! \{1,0,0\}$, empirically leads to poor recovered object edges because the balanced semantics and coarse saliency compete over where the underlying geometry is. Instead, we weight both equally $\lambdaSAFFsem = \lambdaSAFFatt = \{\nicefrac{1}{9},\nicefrac{4}{9},\nicefrac{4}{9} \}$. Even though the coarse layer has smaller weight, it is sufficient to guide the overall result. This balances high resolution edges from fine layers and whole object features from coarse layers while reducing geometry conflicts, and leads to improved features (\cref{fig:feature_improvement}).

Of course, any sliding window must contain an object to extract reliable features for that object. At coarse levels, an object is always in view.
At fine levels, an object is only captured in \emph{some} windows. Objects of interest tend to be near the middle of the frame, meaning that boundary windows at finer pyramid levels contain features that less reliably capture those objects. This can cause spurious connections in clustering. To cope with this, we relatively decrease finer level boundary window weights: We upsample all levels to the finest level, then increase the coarsest level weight towards the frame boundary to $\lambdaSAFFsem = \lambdaSAFFatt = \{\nicefrac{1}{3},\nicefrac{1}{3},\nicefrac{1}{3} \}$. 

%% file: tex/03d_methods_saffcluster.tex
\subsection{Using SAFF for Saliency-aware Clustering}
\label{sec:method_saliencyawareclustering}

We now wish to isolate salient objects. Even in dynamic scenes, relevant objects may not move, so analyzing dynamic elements is insufficient (cf.~\cite{wu2022ddnerf}). One approach predicts segmentation end-to-end~\cite{lin2021video}. However, end-to-end learning requires priors about the scene provided by supervision, and even large-scale pretraining might fail given unseen test scenes. To achieve scene-specific decompositions from per-video semantics, inspired by Amir \etal~\cite{amir2021deep}, we design clustering for spacetime volumes that allows segmenting novel spatio-temporal views. While \semfeatnetwork is trained on images, its features are loosely temporally consistent~\cite{caron2021emerging}.

Some works optimize a representation with a fixed number of clusters, e.g., via slot attention~\cite{locatello2020object} in NeRFs~\cite{stelzner2021decomposing,yu2022unsupervised}. Instead, we cluster using elbow $k$-means, letting us adaptively find the number of clusters after optimization. This is more flexible than baking an anticipated number of slots (sometimes with fixed semantics),
and lets us cluster and segment in novel spatio-temporal viewpoints. 

We demonstrate clustering results in both 3D over time and on rendered volume 2D projections over time. Given the volume reconstruction, we might think that clustering directly in 3D would be better. But, monocular input with narrow baselines makes it challenging to precisely reconstruct geometries: Consider that depth integrated a long a ray can still be accurate even though geometry at specific 3D points may be inaccurate or `fluffy'. As such, we use the 2D volume projection clustering results in 2D comparisons.


\vspace{-0.2cm}
\paragraph{Method}
For 3D over time, we sample points from the \titlename uniformly along input rays ($128\!\times\!H\!\times\!W$) and treat each pixel as a feature point, e.g., semantics are 64\ dim. and saliency is 1\ dim. For volume projection, we render to $N$ input poses ($N\!\times\!H\!\times\!W$), and treat each pixel as a feature point instead.
%
%
In either case, we cluster all feature points together using elbow $k$-means to produce an initial set of separate regions. For each cluster $c$, for each image, we calculate the mean attention of all feature points within the cluster $\bar{\attfeat}_c$. If $\bar{\attfeat}_c > 0.07$, then this cluster is salient for this image. Finally, all feature points vote on saliency: if more than $70\%$ agree, the cluster is salient.


Salient objects may still be split into semantic parts: \eg, in \cref{fig:merge}, the person's head/body are separated. Plus, unwanted background saliency may exist, \eg, input $\hat{\attfeat}^\dagger$ is high for the teal graphic on the wall.
As such, before saliency voting, we merge clusters whose centroids have a cosine similarity $> 0.5$. This reduces the first problem as heads and bodies are similar, and reduces the second problem as merging the graphic cluster into the background reduces its \emph{average} saliency (\cref{fig:merge}).

To extract an object from the 3D volume, we sample 3D points along each input ray, then ascribe the label from the semantically-closest centroid. All clusters not similar to the stored salient clusters are marked as background with zero density. For novel space-time views, we render feature images from the volume, then assign cluster labels to each pixel according to its similarity with stored input view centroids. 

\input{fig/fig_merge/fig_merge.tex}

%% file: fig/fig_merge/fig_merge.tex
%
%

{   

\begin{figure*}[t]
    \centering
    \setlength{\tabcolsep}{1pt}
    
    \newcommand{\figSeqMerge}{fig/fig_merge}

    \newcommand{\graymap}{{0 1}}
    \newcommand{\graymapBrighter}{{0.2 1.2}}
    \newcommand{\rgbcolormap}{{0 1 0 1 0 1}}
    \newcommand{\rgbcolormapBrighter}{{1 0 1 0 1 0}}
    
    \newcommand{\imgw}{0.18\linewidth} 
    \newcommand{\imgwhalf}{0.12\linewidth} 
    
    \newcommand{\imgRGB}[2][0px 0px 100px 0px]{\includegraphics[draft=\isdraft,width=\imgw,trim=#1,clip]{#2}}
    \newcommand{\imgRGBHalf}[2][0px 0px 100px 0px]{\includegraphics[draft=\isdraft,width=\imgwhalf,trim=#1,clip]{#2}}
    
    \newcommand{\imgAtt}[2][0px 0px 100px 0px]{\includegraphics[draft=\isdraft,decodearray={0.2 1.2},width=\imgw,trim=#1,clip]{#2}}
    \newcommand{\imgAttHalf}[2][0px 0px 100px 0px]{\includegraphics[draft=\isdraft,decodearray={0.2 1.2},width=\imgwhalf,trim=#1,clip]{#2}}

    \scriptsize
    \begin{tabular}{c | c c c | c}
    
    
    (i) Input frame &
    (a) Our pyr.~semantic (volume) &
    (b) Semantic clustering (pre-merge) &
    (c) Salient clusters (pre-merge) &
    (iii) Extraction (fixed time)
    \\    

    \imgRGB{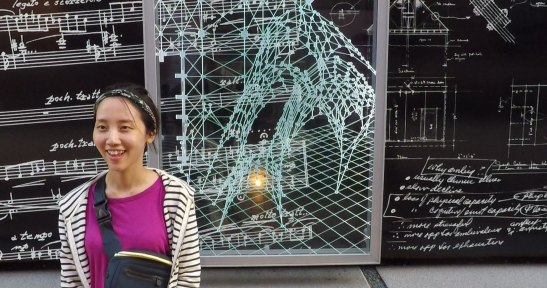} &
    \imgRGB{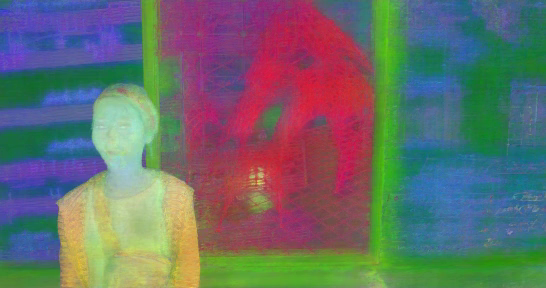} &
    \imgRGB{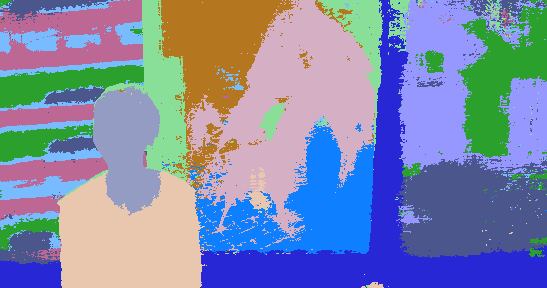} &
    \imgRGB{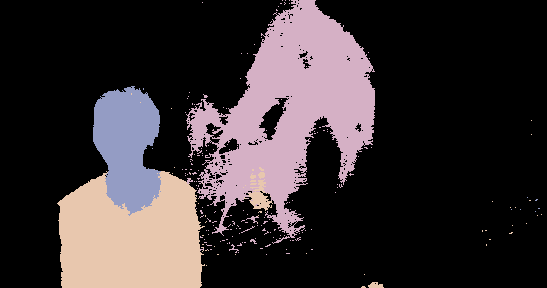} &
    \imgRGB{\figSeqMerge/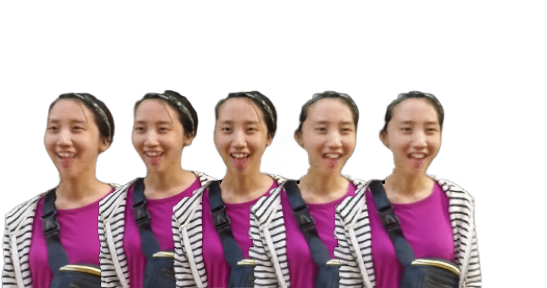}
    \\


    
    (ii) Input semantic/attention &
    (d) Our pyr.~saliency (volume) &
    (e) Our merged clustering &
    (f) Our final foreground &
    (iv) Extraction (fixed view)
    \\

    
    \begin{tikzpicture}
    \node (input-semantics) {
        \imgRGBHalf{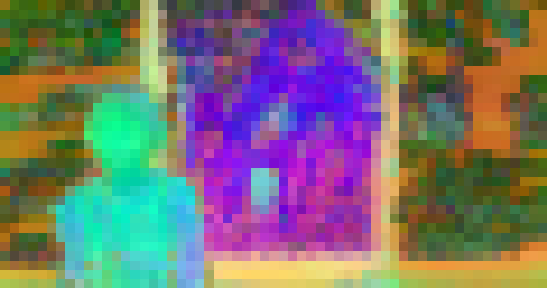} };
    \node (input-saliency) [below right=-1cm and -1.4cm of input-semantics] {
        \imgAttHalf{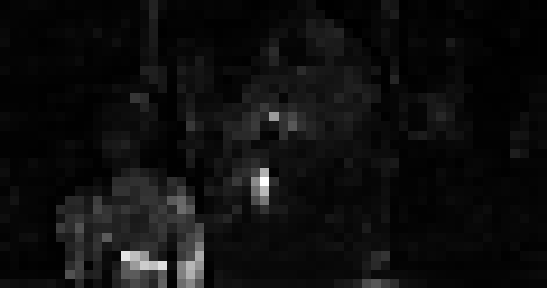} };
    \end{tikzpicture}
    &
    \imgAtt{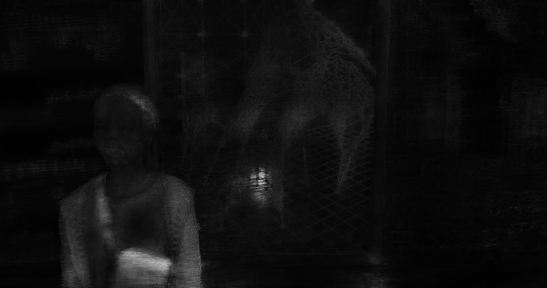} &
    \imgRGB{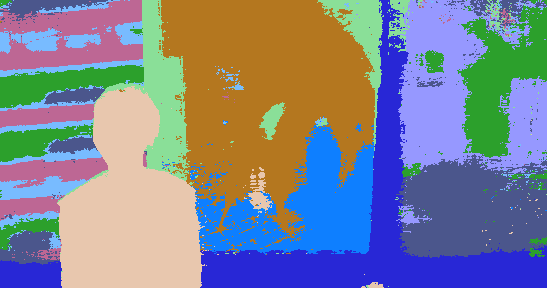} &
    \imgRGB{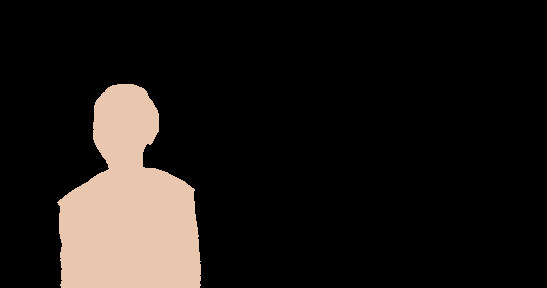} &
    
    \imgRGB{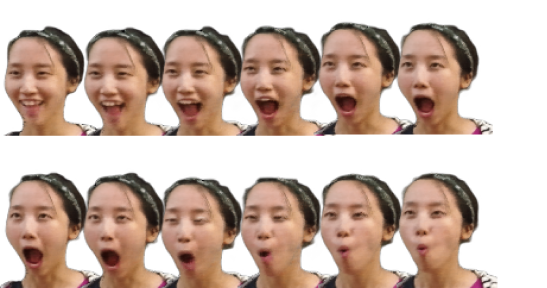}

    \end{tabular}

    \vspace{-0.25cm}
    \caption{
        \textbf{Saliency-aware clustering improves decomposition.} On \seqDynamicFace, the head and body are semantically and saliently different, but are mutually different from the background. This allows us to extract a time-varying 3D field of the object.
    }
	
    \label{fig:merge}
    \vspace{-0.25cm}
\end{figure*}

}

%% file: tex/04_experiments.tex
\vspace{-0.05cm}
\section{Experiments}

We show the impact of adding semantic and saliency features through scene decomposition and foreground experiments. Our website contains supplemental videos.

\vspace{-0.25cm}
\paragraph{Data: NVIDIA \dataname}
This data~\cite{Yoon_2020_CVPR} has 8 sequences of 24 time steps formed from 12 cameras simultaneously capturing video.
We manually annotate object masks for view and time step splits; we will release this data publicly. 
%
We define three data splits per sequence:
\begin{enumerate}[parsep=0pt,topsep=4pt,leftmargin=12pt]
    \item \emph{\datasplitInput}: A monocular camera that moves position for every timestep is simulated from the input sequences; we use Yoon \etal's input sequences~\cite{Yoon_2020_CVPR}. 
    
    \item \emph{\datasplitCamZero}\ (hold out): We fix the camera at position 0 as time plays, requiring novel view and time synthesis. $\{(\mathrm{cam}_0, \mathrm{time}_i), i \in [1, 2, ... 23]\}$.
    \item \emph{\datasplitTimeZero}\ (hold out): We fix time at step 0 as the camera moves, requiring novel view and time synthesis. $\{(\mathrm{cam}_i, \mathrm{time}_0), i \in [1, 2, ... 11]\}$. 
\end{enumerate}

\paragraph{Data: \dycheck}
For single-camera casual monocular data, we select a subset of Gao et al.'s DyCheck dataset~\cite{gao2022dynamic}. We remove scenes without an obvious subject, e.g., where telling foreground from background is hard even for a human. We select \emph{haru-sit}, \emph{space-out}, \emph{spin}, \emph{paper-windmill}, and \emph{mochi-high-five}. We uniformly sample $48$ frames and manually annotate object masks. We take even frames as the input set and odd frames as the hold out set. We use the same hyperparameters between both datasets.


\vspace{-0.3cm}
\paragraph{Metrics}
To assess clustering performance, we use the Adjusted Rand Index (ARI; $[-1, 1]$). This compares the similarity of two assignments without label matching, where random assignment would score $\approx$0. 
For foreground segmentation, we compute IoU (Jaccard), and for RGB quality we use PSNR, SSIM, and LPIPS.

%
%
\input{fig/fig_qualitative_scenedecomp/fig_qualitative_scenedecomp.tex}

\subsection{Comparisons including ablations}


\begin{description}[itemsep=0.4ex, topsep=0.5ex, leftmargin=0cm]
    
    \item[\titlename (ours)] We optimize upon the input split of each scene, and perform clustering to obtain object segmentations. To produce a foreground, we merge all salient objects.
    
     \item[\titlename (3D)] The same as above, but processed in 3D rather than on 2D volume projections.

    \item[--- w/ pyr $\lambdaSAFFatt = \{1,0,0\}$] Pyramid with only coarse saliency (\cref{sec:method_semfeatnetworkprocess}) and balanced semantic weight across levels.

    \item[--- w/o pyr] No pyramid (\cref{sec:method_semfeatnetworkprocess}); we optimize with features and saliency extracted from the input image only.

    \item[--- w/o merge] With pyramid, but we remove cluster merging inside the saliency-aware clustering algorithm.

    \item[--- w/ blend $v$] To compare generic dynamic segmentation to saliency segmentation, we use the static/dynamic weight instead of volume saliency to segment foreground objects. We set every pixel below the 80\% $v$ quantile in each image to be background, or otherwise foreground.
    
    \item[--- w/ post process] We add a step after the saliency-aware clustering to refine edges using a conditional random field (please see supplemental material for details). This gains significantly from the depth estimated via volume reconstruction, producing sharp and detailed edges.

    \textbf{NSFF}~\cite{li2020neural} This method cannot produce semantic clusterings. While saliency and blend weight $v$ have different meanings, if we compare our $v$ to NSFF's, then we can see any impact of a shared backbone with attention heads upon the static/dynamic separation. We disable the supervised motion mask initialization in NSFF as our method does not use such information.

    \textbf{\ddnerf}~\cite{wu2022ddnerf} This method also cannot produce semantic clusterings. Over HyperNeRF~\cite{park2021hypernerf}, it adds a shadow field network and further losses to try to isolate objects into the dynamic NeRF over the separate static one. The paper also compares to NSFF without motion mask initialization.

    \item[\semfeatnetwork (2D)~\cite{amir2021deep}] We ignore the volume and pass 2D semantic and attention features into the clustering algorithm. This cannot apply to novel viewpoints. Instead, we evaluate the approach upon \emph{all} multi-view color images---input and hold-out---whereas other methods must render hold-out views. With our added pyramid processing (\cref{sec:method_semfeatnetworkprocess}).

    \item[--- w/o pyr] No pyramid; upsample to input RGB size.

    \item[ProposeReduce (2D)~\cite{lin2021video}] As a general comparison point, we apply this state-of-the-art 2D video segmentation method. For object segmentation, we use a ProposeReduce network that was pretrained with supervision on YouTube-VIS 2019~\cite{vis2019} for instance segmentation. For foreground segmentation, we use weights pretrained on UVOS~\cite{Caelles_arXiv_2019} intended specifically for unsupervised foreground segmentation. As ProposeReduce is only a 2D method, we provide it with hold-out images for splits with novel views rather than our approach that must render novel views at hold-out poses.
\end{description}


%
%
\subsection{Findings} 




%
%
\input{fig/fig_foreground/fig_foreground_segmentation.tex}

\vspace{-0.15cm}
\paragraph{Dynamic scene decomposition}

We separate the background and each foreground object individually (\cref{tab:scenedecomp}). The baseline 2D \semfeatnetwork method is improved by our pyramid approach. But, being only 2D, this fails to produce a consistent decomposition across novel spacetime views even when given ground truth hold-out images. This shows the value of a volume integration. Next, supervised ProposeReduce produces good results (\cref{fig:qualitative_scenedecomp}), but sometimes misses salient objects or fails to join them behind occluding objects, and only sometimes produces better edges than our method without post-processing as it can oversmooth edges. ProposeReduce also receives ground truth images in hold-out sets.

Instead, our approach must render hold-out views via the volume reconstruction. This produces more consistent segmentations through spacetime manipulations---this is the added value of volume integration through learned saliency and attention heads. Ablated components show the value of our pyramid step, its coarse-saliency-only variant, and the cluster merge and image post processing steps. Qualitatively, we see good detail (\cref{fig:qualitative_scenedecomp}); post-processing additionally improves edge quality and removes small unwanted regions.

\vspace{-0.15cm}
\paragraph{Oracle saliency}
We also include an oracle experiment where saliency voting is replaced by cluster selection based on ground truth masks. This experiment tells us what part of the performance gap lies with saliency itself, and what remains due to volume integration and cluster boundary issues. With oracle clusters, our decomposition performance is 0.8 ARI (\cref{tab:scenedecomp}) even in hold-out views. This shows that our existing cluster boundaries are accurate, and that accurate saliency for object selection is the larger remaining problem. 

\input{tab/tab_scenedecomp.tex}

\input{tab/tab_foreground_segmentation.tex}

\vspace{-0.15cm}
\paragraph{Foreground segmentation}

We simplify the problem and consider all objects as simply `foreground' to compare to methods that do not produce per object masks. Here, the same trend continues (\cref{tab:foregroundsegmentation}). We note more subtle improvements to static/dynamic blending weights when adding our additional feature heads to the backbone MLP, and overall show that adding top-down information helps produce more useful object masks.
Qualitative results show whole objects in the foreground rather than broad regions of possible dynamics (NSFF) or broken objects (\ddnerf; \cref{fig:qualitative_foreground_segmentation}).

\input{tab/tab_dycheck.tex}
\input{fig/fig_dycheck/fig_dycheck_single.tex}

\vspace{-0.25cm}
\paragraph{DyCheck evaluation}
DyCheck often has close-up objects and, even with our selected sequences, saliency struggles to find foreground objects. Thus, we use oracle saliency. 
\semfeatnetwork and ProposeReduce are given test views while \titlename must render them. Quantitative (\cref{tab:dycheck}) and qualitative (\cref{fig:dycheck}) experiments show similar trends as before: ProposeReduce is good but may miss objects and fine details; \titlename may produce finer details and is more geometrically consistent. 


%% file: fig/fig_qualitative_scenedecomp/fig_qualitative_scenedecomp.tex

{   

\begin{figure*}[t]
    \centering
    \setlength{\tabcolsep}{1pt}
    
    \newcommand{\figSeqBalloonNoticeBoard}{fig/fig_qualitative_scenedecomp/balloonnoticeboard}
    \newcommand{\figSeqBalloonWall}{fig/fig_qualitative_scenedecomp/balloonwall}
    \newcommand{\figSeqDynamicFace}{fig/fig_qualitative_scenedecomp/dynamicface}
    \newcommand{\figSeqJumping}{fig/fig_qualitative_scenedecomp/jumping}
    \newcommand{\figSeqPlayground}{fig/fig_qualitative_scenedecomp/playground}
    \newcommand{\figSeqSkating}{fig/fig_qualitative_scenedecomp/skating}
    \newcommand{\figSeqTruck}{fig/fig_qualitative_scenedecomp/truck}
    \newcommand{\figSeqUmbrella}{fig/fig_qualitative_scenedecomp/umbrella}

    \newcommand{\rgbcolormapBrighter}{{0.2 1.2 0.2 1.2 0.2 1.2}}
    \newcommand{\rgbcolormap}{{0 1 0 1 0 1}}

    \newcommand{\imgw}{0.18\linewidth}
    \newcommand{\imgh}{0.01\linewidth}
    
    \newcommand{\imgRGB}[2][0px 0px 0px 0px]{\includegraphics[draft=\isdraft,width=\imgw,trim=#1,clip]{#2}}

    
    \newcommand{\zoomRGB}[4]{
        \begin{tikzpicture}[
    		image/.style={inner sep=0pt, outer sep=0pt},
    		collabel/.style={above=9pt, anchor=north, inner ysep=0pt, align=center}, 
    		rowlabel/.style={left=9pt, rotate=90, anchor=north, inner ysep=0pt, scale=0.8, align=center},
    		subcaption/.style={inner xsep=0.75mm, inner ysep=0.75mm, below right},
    		arrow/.style={-{Latex[length=2.5mm,width=4mm]}, line width=2mm},
    		spy using outlines={rectangle, size=1.25cm, magnification=3, connect spies, ultra thick, every spy on node/.append style={thick}},
    		style1/.style={cyan!90!black,thick},
    		style2/.style={orange!90!black},
    		style3/.style={blue!90!black},
    		style4/.style={green!90!black},
    		style5/.style={white},
    		style6/.style={black},
        ]
        
        \node [image] (#1) {\imgRGB{#2}};
        \spy[style5] on ($(#1.center)-#3$) in node (crop-#1) [anchor=#4] at (#1.#4);
        
        \end{tikzpicture}
    }

    
    \begin{tabular}{c c@{\hspace{0.2mm}} c@{\hspace{0.2mm}} c@{\hspace{0.2mm}} c@{\hspace{0.2mm}} c}
    
    
    & 
    (a) Input &
    (b) \semfeatnetwork (2D) & 
    (c) ProposeReduce~\cite{lin2021video} &
    (d) \titlename~(Ours; input) &
    (e) \titlename~(Ours; novel) \\
    
    \rotatebox{90}{\footnotesize \seqDynamicFace} 

    &
    \zoomRGB{balloonw-rgb-input}{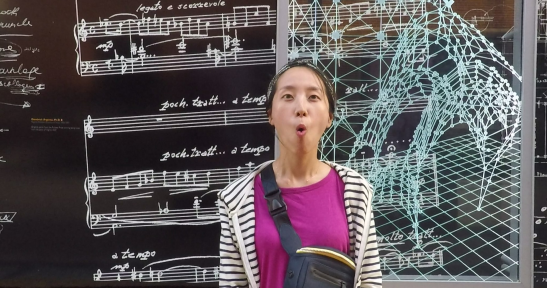}{(0.1,0.05)}{north west} &
    
    \zoomRGB{balloonw-clustering-dinovit2d}{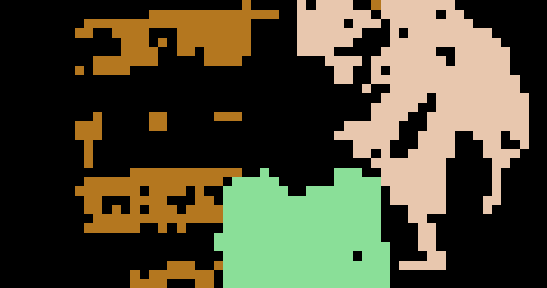}{(0.1,0.05)}{north west} &
    
    \zoomRGB{balloonw-clustering-propred}{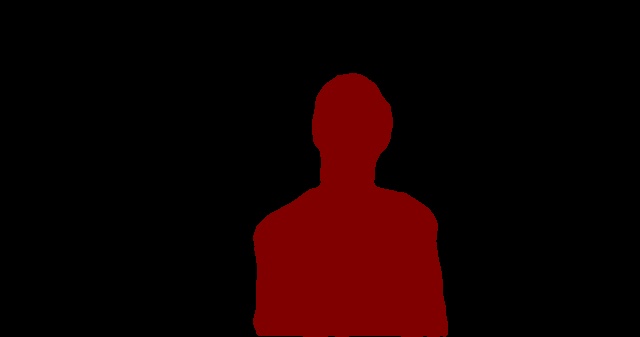}{(0.1,0.05)}{north west} &
    
    \zoomRGB{balloonw-clustering-oursiv}{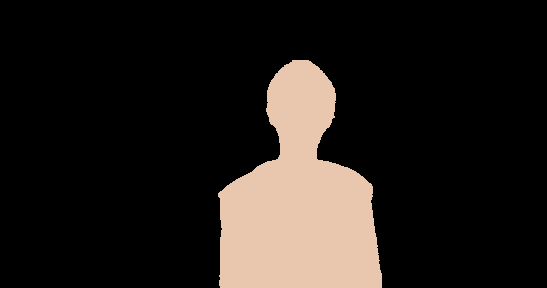}{(0.1,0.05)}{north west} & 

    \zoomRGB{balloonw-clustering-oursnv}{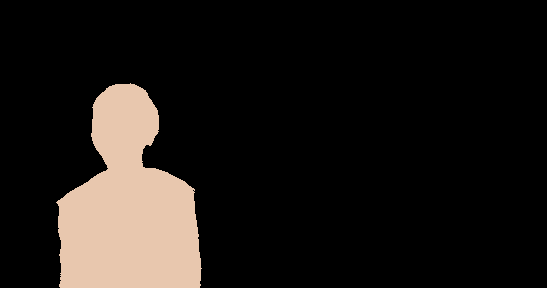}{(1.10,0.15)}{north east} 
    \\[-1pt]

    \rotatebox{90}{\footnotesize \seqBalloonNoticeBoard}

    &
    \zoomRGB{balloonnb-rgb-input}{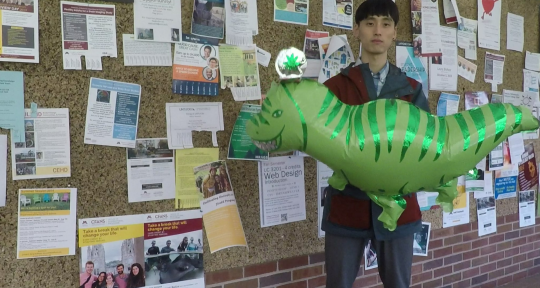}{(-0.7,0.35)}{south west} &
    
    \zoomRGB{balloonnb-clustering-dinovit2d}{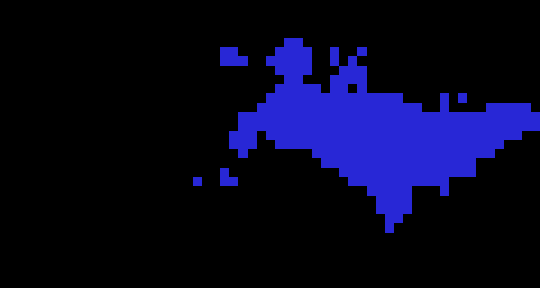}{(-0.7,0.35)}{south west} &
    
    \zoomRGB{balloonnb-clustering-propred}{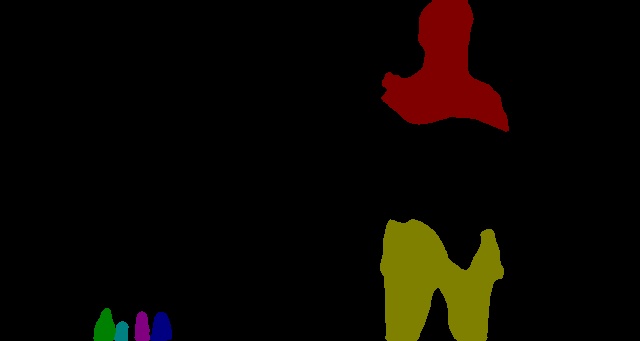}{(-0.7,0.35)}{south west} &
    
    \zoomRGB{balloonnb-clustering-oursiv}{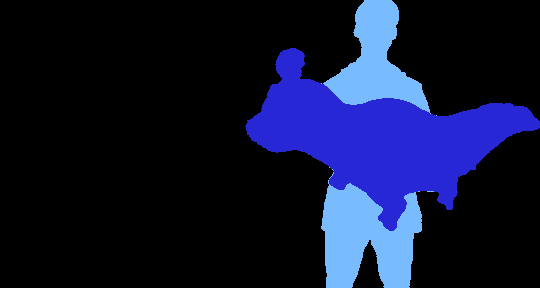}{(-0.7,0.35)}{south west} &

    \zoomRGB{balloonnb-clustering-oursnv}{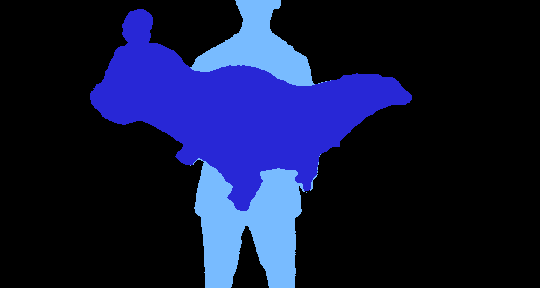}{(0.23,0.25)}{south east} 
    \\[-1pt]

    \rotatebox{90}{\footnotesize \hspace{2.5mm} \seqUmbrella} 

    &
    \zoomRGB{balloonw-rgb-input}{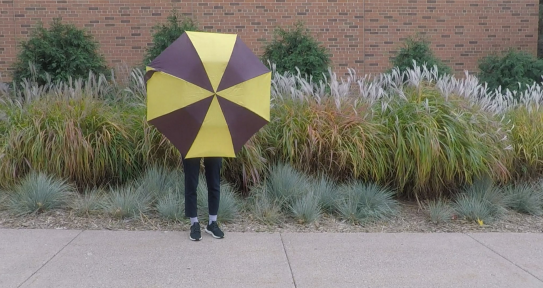}{(0.15,-0.5)}{north east} &
    
    \zoomRGB{balloonw-clustering-dinovit2d}{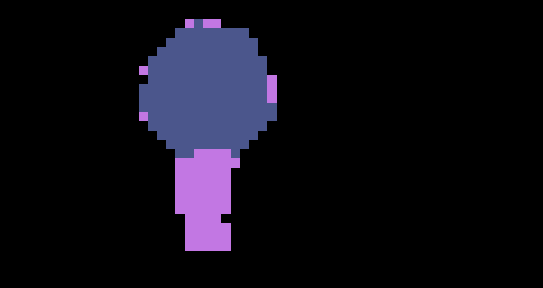}{(0.15,-0.5)}{north east} &
    
    \zoomRGB{balloonw-clustering-propred}{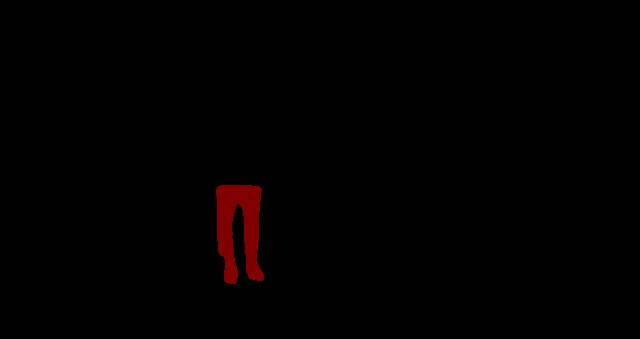}{(0.15,-0.5)}{north east} &
    
    \zoomRGB{balloonw-clustering-oursiv}{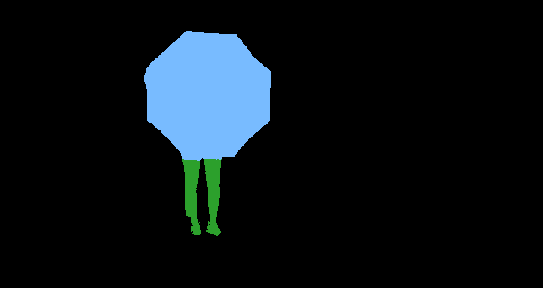}{(0.15,-0.5)}{north east} & 

    \zoomRGB{balloonw-clustering-oursnv}{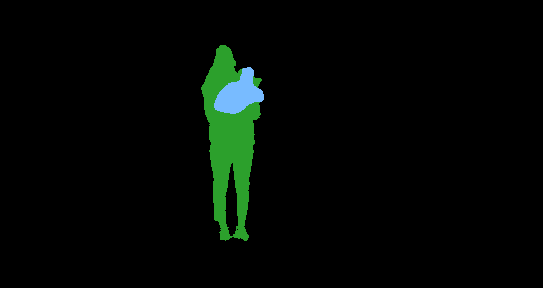}{(0.15,-0.5)}{north east} 
    \\[-1pt]



    \end{tabular}

    \vspace{-0.25cm}
    \caption{
        \textbf{\titlename object segmentations show balanced quality while recovering a volumetric scene representation (e).}
        Basic \semfeatnetwork produces low-quality segmentations and misses objects. A state-of-the-art 2D video learning method~\cite{lin2021video} sometimes has edge detail (\seqUmbrella, legs) but othertimes misses detail and objects (\seqBalloonNoticeBoard). 
    }
    \label{fig:qualitative_scenedecomp}
\end{figure*}

}

%% file: fig/fig_foreground/fig_foreground_segmentation.tex

{   

\begin{figure*}[t]
    \centering
    \setlength{\tabcolsep}{1pt}
    
    \newcommand{\figSeqJumping}{fig/fig_foreground/jumping}
    \newcommand{\figSeqTruck}{fig/fig_foreground/truck}
    \newcommand{\figSeqBalloon}{fig/fig_foreground/balloon}

    \newcommand{\rgbcolormapBrighter}{{0.2 1.2 0.2 1.2 0.2 1.2}}
    \newcommand{\rgbcolormap}{{0 1 0 1 0 1}}

    \newcommand{\imgw}{0.18\linewidth}
    
    \newcommand{\imgRGB}[2][0px 0px 0px 0px]{\includegraphics[draft=\isdraft,decodearray=\rgbcolormap,width=\imgw,trim=#1,clip]{#2}}


    \vspace{-0.4cm} 
    \begin{tabular}{c c c c c c}
    
    & 
    (a) Input &
    (b) NSFF~\cite{li2020neural} blend $v$ &
    (c) \ddnerf~\cite{wu2022ddnerf} blend $v$ &
    (d) \titlename blend $v$ & 
    (e) \titlename foregrounds \\

           
    \rotatebox{90}{{\footnotesize \hspace{3mm} \seqJumping}}
    &
    \includegraphics[width=\imgw]{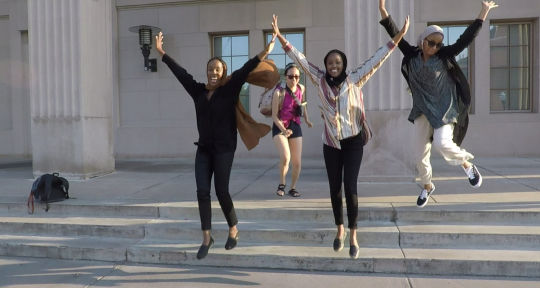} &
    \includegraphics[width=\imgw]{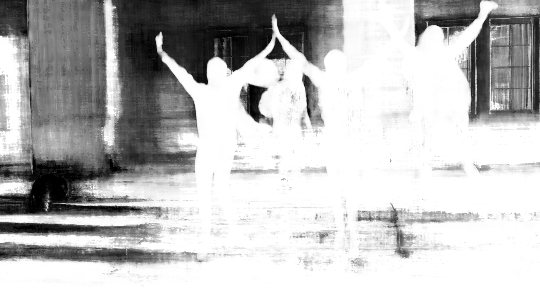} &
    \includegraphics[width=\imgw]{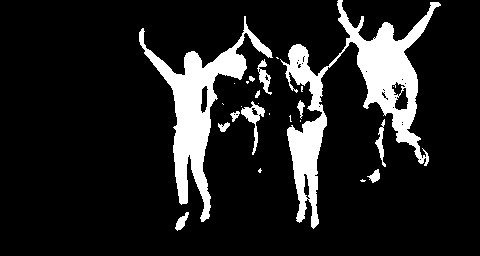} &
    \includegraphics[width=\imgw]{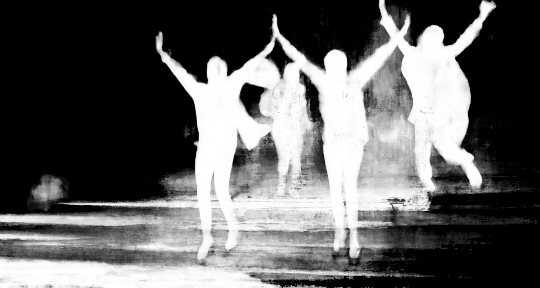} &
    \includegraphics[width=\imgw]{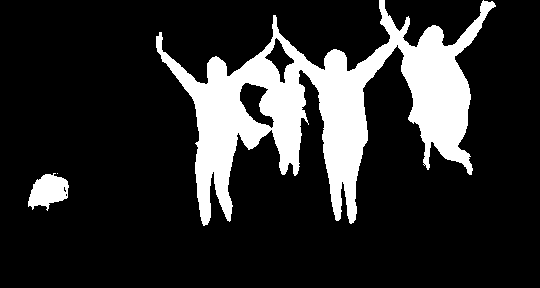}
    \\

     
    
     
    \rotatebox{90}{\footnotesize \seqBalloonNoticeBoard}
    &
    \includegraphics[width=\imgw]{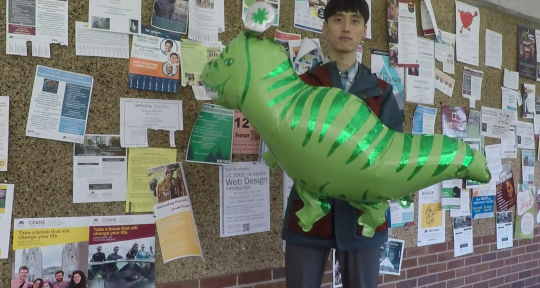} &
    \includegraphics[width=\imgw]{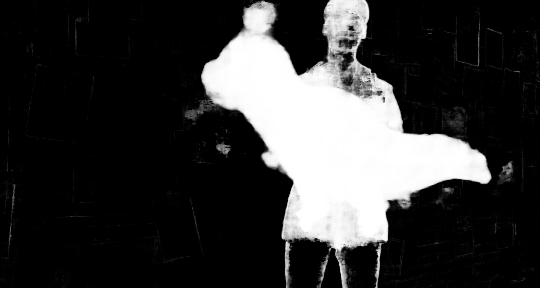} &
    \includegraphics[width=\imgw]{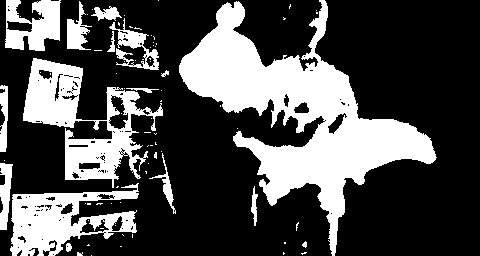} &
    \includegraphics[width=\imgw]{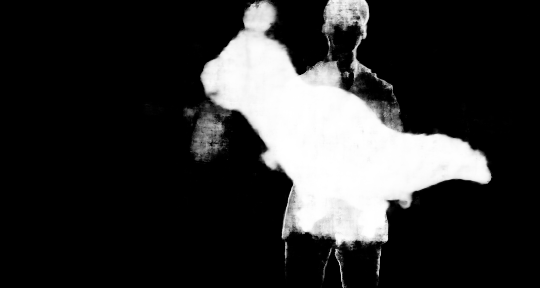} &
    \includegraphics[width=\imgw]{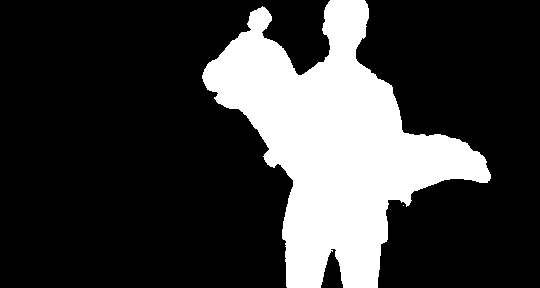}



    \end{tabular}

    \vspace{-0.3cm}
    \caption{
        \textbf{Saliency improves foreground segmentation.} Static/dynamic separations are not foreground segmentations, leading to limited use of dynamic NeRF models for downstream tasks. Minor improvements to dynamic blending weight $v$ are seen in some sequences (\seqJumping) by adding the saliency head to the shared backbone.
    }
	
    \label{fig:qualitative_foreground_segmentation}
    \vspace{-0.35cm}
\end{figure*}

}

%% file: tab/tab_scenedecomp.tex
\begin{table}[t]
    \centering
    \caption{
        \textbf{Spacetime volume integration improves dynamic scene decomposition.} Pyramid construction and cluster merging help quantitatively, and ours is comparable to SOTA supervised 2D video segmentation network ProposeReduce. Metric: Adjusted Rand Index ($[-1,1]$, higher is better).
        \label{tab:scenedecomp}
        }
    \vspace{-0.35cm}
    \resizebox{\linewidth}{!}{
    \begin{tabular}{l r r r}
    
    \toprule
    {} & \datasplitInput & \datasplitCamZero & \datasplitTimeZero \\
    \midrule 
    ProposeReduce (2D) &  0.725 & 0.736   & 0.742  \\
    
    \semfeatnetwork (2D)                       &   0.501 & 0.495 & 0.321 \\
    \hspace{0.2cm} w/o pyr $\hat{\semfeat},\hat{\attfeat}$   &   0.470 & 0.464 & 0.346 \\
    
    \midrule
    \titlename (3D)                 &  0.594   &  0.578   & 0.566 \\
    \titlename (ours)                               &  0.653   &  0.634   & 0.625  \\
    \hspace{0.2cm} w/ pyr $\lambdaSAFFatt = \{1,0,0\}$    &  0.620            &  0.598            & 0.592  \\
    
    \hspace{0.2cm} w/o pyr $\hat{\semfeat},\hat{\attfeat}$  &  0.545            &  0.532            & 0.521  \\
    \hspace{0.2cm} w/o merge cluster                &  0.593        & 0.574         & 0.563 \\
    \hspace{0.2cm} w/ post process                  &  0.759        & 0.733         & 0.735 \\
    \hspace{0.2cm} w/ oracle                        &  0.834        & 0.806         & 0.800 \\
    \hspace{0.2cm} w/ oracle + post process         &  0.922        & 0.890         & 0.880 \\
    \bottomrule
    \end{tabular}
    
    }
    \vspace{-0.5cm}
\end{table}

%% file: tab/tab_foreground_segmentation.tex
\begin{table}[t]
    \centering
    
    \caption{
        \textbf{Saliency improves foreground segmentation.}
        Adding saliency also slightly aids how much static/dynamic blend weight $v$ represents the foreground (cf.~NSFF blend $v$ to SAFF's).
        Here, ProposeReduce uses unsupervised training and data~\cite{Caelles_arXiv_2019}.
        Metric: IoU/Jaccard ($[0,1]$, higher is better).
        \label{tab:foregroundsegmentation}
    }
    \vspace{-0.2cm}
    \resizebox{0.92\linewidth}{!}{
    \begin{tabular}{l r r r}
    
    \toprule
    {} & \datasplitInput & \datasplitCamZero & \datasplitTimeZero \\
    \midrule
    ProposeReduce (2D)          &  
    0.609 & 0.464   & 0.591  \\
    DINO-ViT (2D)  &  0.381 & 0.382  & 0.357  \\
    \midrule
    NSFF --- blend $v$          & 0.322 & 0.309 & 0.268 \\
    \ddnerf --- blend $v$       &  0.470 & 0.334 & 0.269 \\ 

    \titlename~--- saliency     &  0.609  &  0.589  & 0.572  \\
    \hspace{0.2cm} --- blend $v$    &  0.388 & 0.380  & 0.329 \\
    \hspace{0.2cm} --- post process $v$    &  0.720 & 0.694  & 0.679 \\
    \bottomrule
    \end{tabular}
    }
    \vspace{-0.1cm}
\end{table}

%% file: tab/tab_dycheck.tex
\begin{table}[t]
\centering
\caption{
        \textbf{SAFF semantics generalize to \dycheck} Oracle saliency. ProposeReduce is given ground truth test frames, while SAFF must render them. \titlename is comparable to SOTA supervised 2D video segmentation network ProposeReduce. 
        Metric: Adjusted Rand Index ($[-1,1]$, higher is better);  IoU/Jaccard index ($[-1,1]$, higher is better).
        \label{tab:dycheck}
        }
\vspace{-0.5cm}
\resizebox{\linewidth}{!}{

\fontsize{8pt}{8pt}\selectfont
\begin{tabular}{l r r r r}
\toprule
\multirow{2}{*}{~}  & \multicolumn{2}{c}{ARI}               & \multicolumn{2}{c}{IoU} \\
\cmidrule{2-5}
                    & \datasplitInput   & \datasplitTest    & \datasplitInput   & \datasplitTest \\
\midrule
ProposeReduce (2D)  & 0.761             & 0.762             & 0.761             & 0.762 \\
\semfeatnetwork (2D)& 0.801             & 0.800             & 0.797             & 0.797 \\
\titlename (3D)     & 0.902             & 0.820             & 0.910             & 0.812 \\
\bottomrule
\end{tabular}

}
\vspace{-0.25cm}
\end{table}

%% file: fig/fig_dycheck/fig_dycheck_single.tex

{   

\begin{figure}[t]
    \centering
    \setlength{\tabcolsep}{1pt}
    
    \newcommand{\figSeqHaruSit}{fig/fig_dycheck/haru-sit}
    \newcommand{\figSeqMochiHighFive}{fig/fig_dycheck/mochi-high-five}
    \newcommand{\figSeqSpaceOut}{fig/fig_dycheck/space-out}
    \newcommand{\figSeqSpin}{fig/fig_dycheck/spin}

    \newcommand{\rgbcolormapBrighter}{{0.2 1.2 0.2 1.2 0.2 1.2}}
    \newcommand{\rgbcolormap}{{0 1 0 1 0 1}}

    \newcommand{\imgw}{0.24\linewidth}
    
    \newcommand{\imgRGB}[2][0px 0px 0px 0px]{\includegraphics[draft=\isdraft,width=\imgw,trim=#1,clip]{#2}}

    
    \newcommand{\zoomRGB}[4]{
        \begin{tikzpicture}[
    		image/.style={inner sep=0pt, outer sep=0pt},
    		collabel/.style={above=9pt, anchor=north, inner ysep=0pt, align=center}, 
    		rowlabel/.style={left=9pt, rotate=90, anchor=north, inner ysep=0pt, scale=0.8, align=center},
    		subcaption/.style={inner xsep=0.75mm, inner ysep=0.75mm, below right},
    		arrow/.style={-{Latex[length=2.5mm,width=4mm]}, line width=2mm},
    		spy using outlines={rectangle, size=0.7cm, magnification=2.5, connect spies, ultra thick, every spy on node/.append style={thick}},
    		style1/.style={cyan!90!black,thick},
    		style2/.style={orange!90!black},
    		style3/.style={blue!90!black},
    		style4/.style={green!90!black},
    		style5/.style={white},
    		style6/.style={black},
        ]
        
        \node [image] (#1) {\imgRGB{#2}};
        \spy[style5] on ($(#1.center)-#3$) in node (crop-#1) [anchor=#4] at (#1.#4);
        
        \end{tikzpicture}
    }

    \scriptsize
    
    \begin{tabular}{c c c c}
    
    Input &
    ProposeReduce~\cite{lin2021video} &
    \titlename~(Ours; input) &
    \titlename~(Ours; novel) \\
    

    \zoomRGB{balloonw-rgb-input}{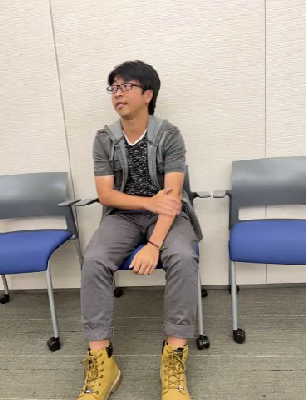}{(0.4,0.95)}{north west} &
    
    
    \zoomRGB{balloonw-clustering-propred}{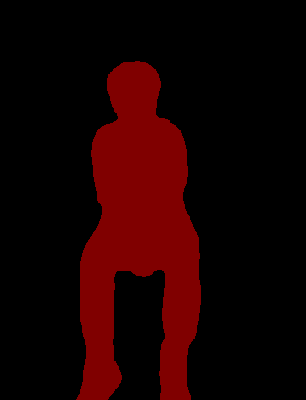}{(0.4,0.95)}{north west} &
    
    \zoomRGB{balloonw-clustering-oursiv}{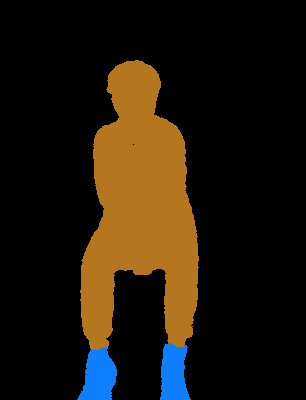}{(0.4,0.95)}{north west} & 

    \zoomRGB{balloonw-clustering-oursnv}{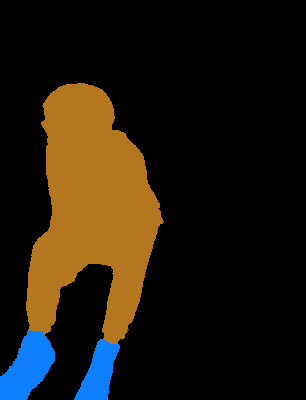}{(0.3, 0.9)}{north east} 
    \\
%
%
%
%
%
    \zoomRGB{balloonw-rgb-input}{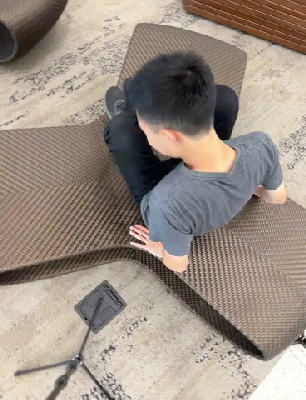}{(0.1,0.25)}{north west} &
    
    
    \zoomRGB{balloonw-clustering-propred}{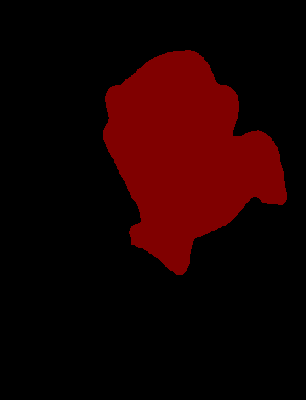}{(0.1,0.25)}{north west} &
    
    \zoomRGB{balloonw-clustering-oursiv}{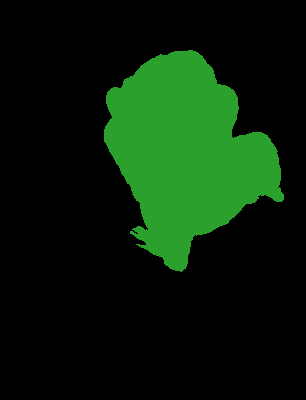}{(0.1,0.25)}{north west} & 

    \zoomRGB{balloonw-clustering-oursnv}{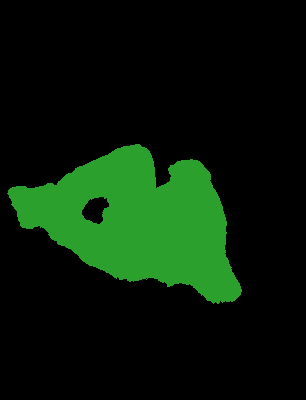}{(0.4,0.05)}{north east} 
    \\[-1pt]

    \end{tabular}

    \vspace{-0.25cm}
    \caption{
        \textbf{\titlename can apply to \dycheck.}
        \titlename is comparable to supervised large-scale 2D video learning~\cite{lin2021video}. Some fine details are improved (bottom, fingers) and novel views maintain their geometry (bottom, gap between legs occluded by head at input timestep).
    }
    \label{fig:dycheck}
    \vspace{-0.25cm}
\end{figure}
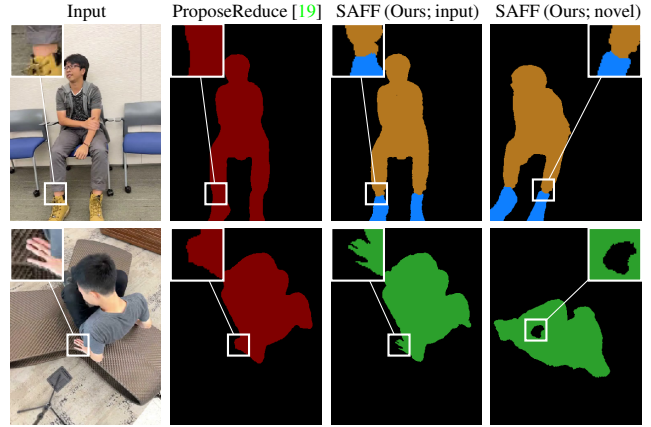

}

%% file: tex/05_discussion.tex
\section{Discussion and Limitations}

\input{fig/fig_limitations/fig_limitations}
\vspace{-0.18cm}

\semfeatnetwork features are not instance-aware (\cref{fig:limitations}). This is in contrast to object-centric learning approaches that aim to identify individual objects. To represent these different approaches, we compare to a result from slot-attention based SAVi++~\cite{elsayed2022savipp}. This method trains on thousands of supervised MOVi~\cite{greff2021kubric} sequences with per-object masks, whereas we use generic pre-trained features and gain better edges from volume integration. Combining these two approaches could give accurate instance-level scene objects.

\semfeatnetwork saliency may attend to unwanted regions. In \Cref{fig:limitations}d--e, the static pillars could be isolated using scene flow. But, often our desired subjects do not move (cf.~people in \seqUmbrella or \seqBalloonNoticeBoard). For tasks or data that can assume that salient objects are dynamic, we use \titlename's 4D scene reconstruction to reject static-but-salient objects by merging clusters via scene flow: First, we project $\flowscene$ over each timestep into each input camera pose---this simulates optical flow with a static camera. Clusters are marked as salient per image if mean flow magnitude per cluster $\bar{|\mathbf{p}|}>0.07$ \emph{and} mean attention $\bar{\attfeat}_c > 0.07$. Finally, as before, a cluster is globally salient if 70\% of images agree 
(\cref{fig:limitations}f).


%% file: fig/fig_limitations/fig_limitations.tex
{
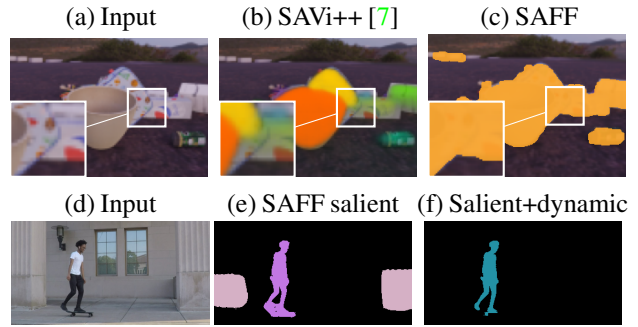
\begin{figure}[t]
    \centering
    \setlength{\tabcolsep}{1pt}
    
    \newcommand{\figSeq}{fig/fig_limitations}
    \newcommand{\moviSeq}{fig/fig_limitations/movi_basket}

    \newcommand{\rgbcolormapBrighter}{{0.2 1.2 0.2 1.2 0.2 1.2}}
    \newcommand{\rgbcolormap}{{0 1 0 1 0 1}}

    \newcommand{\imgw}{0.315\linewidth}
    \newcommand{\twoimgw}{0.5\linewidth}
    \newcommand{\tinyimgw}{0.2\linewidth}
    
    \newcommand{\imgRGB}[2][0px 0px 0px 0px]{\includegraphics[draft=\isdraft,width=\imgw,trim=#1,clip]{#2}}
    
     \newcommand{\zoomRGB}[5]{
        \begin{tikzpicture}[
    		image/.style={inner sep=0pt, outer sep=0pt},
    		collabel/.style={above=9pt, anchor=north, inner ysep=0pt, align=center}, 
    		rowlabel/.style={left=9pt, rotate=90, anchor=north, inner ysep=0pt, scale=0.8, align=center},
    		subcaption/.style={inner xsep=0.75mm, inner ysep=0.75mm, below right},
    		arrow/.style={-{Latex[length=2.5mm,width=4mm]}, line width=2mm},
    		spy using outlines={rectangle, size=1cm, magnification=2, connect spies, ultra thick, every spy on node/.append style={thick}},
    		style1/.style={cyan!90!black,thick},
    		style2/.style={orange!90!black},
    		style3/.style={blue!90!black},
    		style4/.style={green!90!black},
    		style5/.style={white},
    		style6/.style={black},
        ]
        
        \node [image] (#1) {\includegraphics[width=\imgw, trim=#5,clip]{#2}};
        \spy[style5] on ($(#1.center)-#3$) in node (crop-#1) [anchor=#4] at (#1.#4);
        
        \end{tikzpicture}
    }
    
    
    \begin{tabular}{c c c}
     
    (a) Input &
    (b) SAVi++~\cite{elsayed2022savipp} &
    (c) \titlename
    \\
     
    \zoomRGB{movi-basket-input}{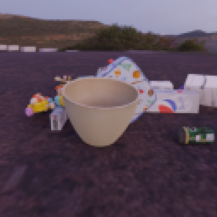}{(-0.5,0)}{south west}{0px 40px 0px 25px} &
    \zoomRGB{movi-basket-savi}{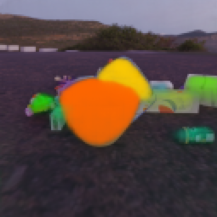}{(-0.5,0)}{south west}{0px 40px 0px 25px} &
    \zoomRGB{movi-basket-ours}{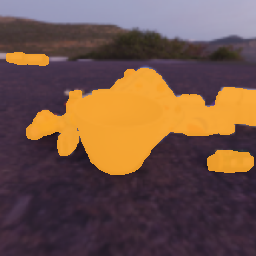}{(-0.5,0)}{south west}{0px 42px 0px 30px}

    \end{tabular}

    \begin{tabular}{c c c}
    

    

    (d) Input &
    (e) \titlename salient &
    (f) Salient+dynamic
    \\


    
   
    \includegraphics[width=\imgw,trim=80px 50px 100px 50px,clip]{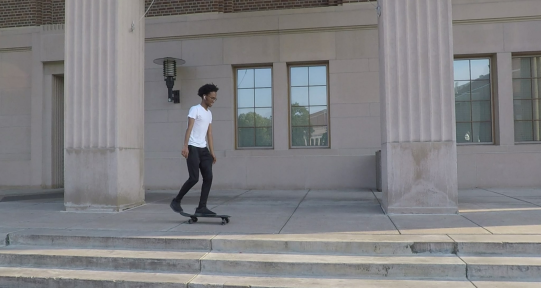} &
    \includegraphics[width=\imgw,trim=80px 50px 100px 50px,clip]{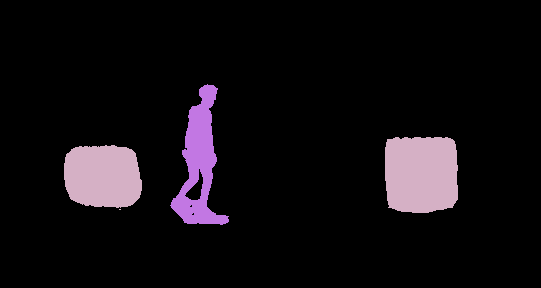} &
    
    \includegraphics[width=\imgw,trim=80px 50px 100px 50px,clip]{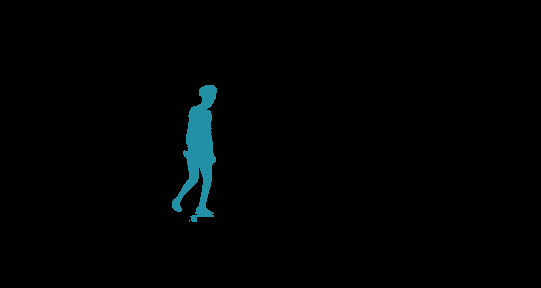}

    \end{tabular}

    \vspace{-0.25cm}
    \caption{
        \textbf{Limitations.} 
        \semfeatnetwork is not instance-aware, causing unwanted grouping (c). Unwanted static objects may be salient; assuming salient objects are dynamic fixes this (f).
    }
	
    \label{fig:limitations}
    \vspace{-0.5cm}
\end{figure}

}